\documentclass[10pt,twocolumn,letterpaper]{article}

 \usepackage{cvpr}              % To produce the CAMERA-READY version
\usepackage[dvipsnames]{xcolor}

\definecolor{cvprblue}{rgb}{0.21,0.49,0.74}
\usepackage[pagebackref,breaklinks,colorlinks,citecolor=cvprblue]{hyperref}

\usepackage{amsmath,amsfonts,bm}

\def\eqref#1{equation~\ref{#1}}
\def\1{\bm{1}}

\def\vx{{\bm{x}}}

\DeclareMathAlphabet{\mathsfit}{\encodingdefault}{\sfdefault}{m}{sl}
\SetMathAlphabet{\mathsfit}{bold}{\encodingdefault}{\sfdefault}{bx}{n}

\newcommand{\bh}{\mathbf{h}}

\newcommand{\bc}{\mathbf{c}}

\newcommand{\bv}{\mathbf{v}}

\newcommand{\bx}{\mathbf{x}}

\newcommand{\bz}{\mathbf{z}}

\usepackage{url}

\usepackage{graphicx}
\usepackage{amsmath}
\usepackage{amssymb}
\usepackage{booktabs}
\usepackage{times}
\usepackage{microtype}
\usepackage{epsfig}
\usepackage{caption}
\usepackage{xcolor}
\usepackage{float}
\usepackage{placeins}
\usepackage{color, colortbl}
\usepackage{stfloats}
\usepackage{enumitem}
\usepackage{tabularx}
\usepackage{xstring}
\usepackage{multirow}
\usepackage{xspace}
\usepackage{url}
\usepackage{subcaption}
\usepackage{arydshln}
\usepackage{xcolor}
\usepackage{bbm}
\usepackage[hang,flushmargin]{footmisc}
\usepackage{pifont}%

\usepackage{wrapfig}
\usepackage{lipsum}
\usepackage{comment}
\usepackage{algorithm}
\usepackage{algorithmic}
\usepackage{listings}

\newcommand{\et}[2]{${#1}^{\pm{#2}}$}

\usepackage{xcolor}   
\definecolor{mydarkblue}{rgb}{0,0.08,0.45}
\usepackage[capitalize]{cleveref}
\crefname{section}{Sec.}{Secs.}
\Crefname{section}{Section}{Sections}
\Crefname{table}{Table}{Tables}
\crefname{table}{Tab.}{Tabs.}

\newcommand{\cs}[1]{\textcolor{orange}{CS: #1}}

\def\paperID{12564} % *** Enter the Paper ID here
\def\confName{CVPR}
\def\confYear{2024}

\title{Motion Flow Matching for  Human Motion Synthesis and Editing}

\author
{
Vincent Tao Hu$^{1,2}\thanks{Partial works are done in UvA. \textsuperscript{$\dagger$} denotes co-supervision.}$,
Wenzhe Yin$^{2}$,
Pingchuan Ma$^{1}$, 
Yunlu Chen$^{3}$,  
Basura Fernando$^{4}$,  \\
Yuki M Asano$^{2}$,   Efstratios Gavves$^{2}$,   Pascal Mettes$^{2}$,   Björn Ommer$^{1,}$\textsuperscript{$\dagger$},  Cees G. M. Snoek$^{2,}$\textsuperscript{$\dagger$}, \\
$^1$CompVis Group, LMU Munich, $^2$University of Amsterdam, $^3$CMU  , $^4$A*STAR \\
}

\begin{document}
\maketitle
\begin{abstract}
Human motion synthesis is a fundamental task in computer animation. Recent methods based on diffusion models or GPT structure demonstrate commendable performance but exhibit drawbacks in terms of slow sampling speeds and error accumulation. In this paper, we propose \emph{Motion Flow Matching}, a novel generative model designed for human motion generation featuring efficient sampling and effectiveness in motion editing applications. Our method reduces the sampling complexity from thousand steps in previous diffusion models to just ten steps, while achieving comparable performance in text-to-motion and action-to-motion generation benchmarks.  Noticeably, our approach establishes a new state-of-the-art Fréchet Inception Distance on the KIT-ML dataset. What is more, we tailor a straightforward motion editing paradigm named \emph{sampling trajectory rewriting} leveraging the ODE-style generative models and apply it to various editing scenarios including motion prediction, motion in-between prediction, motion interpolation, and upper-body editing. Our code will be released. %We provide our motion generation visualization video in the supplementary files. 

\end{abstract}    

\section{Introduction}

\begin{figure}
  \centering
  % include second image
\includegraphics[width=.98\linewidth]{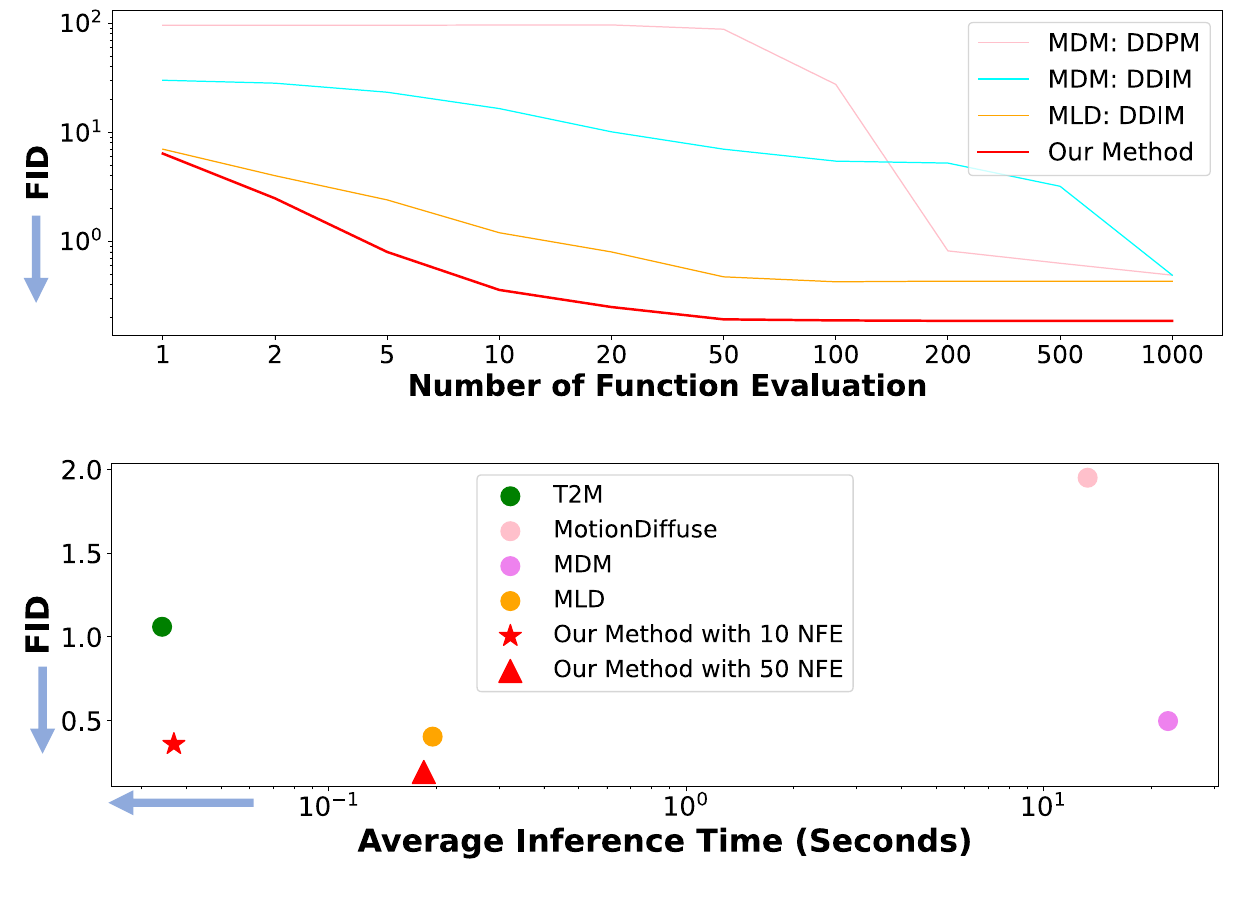}
  %\caption{\textbf{FID \textit{v.s.} NFE on ImageNet100.}}
\caption{
%\cs{This caption is a bit cryptic. The image quality is also below-par. Why needed here? Can also go to experiments, so you have room for a more compelling figure 1?}
\textbf{FID vs. Number of Function Evaluation and FID vs. Average Inference Time} on the KIT-ML dataset. Our method outperforms the baselines, achieving superior FID scores while maintaining faster sampling times. Please note that some axes in the plots are log-scaled for better comparison.}
%\label{fig:guidance_strength}
\label{fig:teaser}
\end{figure}

Human motion generation~\citep{guo2022generating,zhu2023human_motionsurvey23} constitutes a foundational task in computer animation with diverse applications spanning computer graphics, human-computer interaction, and robotics. In contrast to unconditional motion generation~\citep{petrovich2021action,raab2023modi}, recent endeavors have focused on introducing different conditions for enhanced controllability, such as action name~\citep{tevet2022human_mdm}, text~\citep{chen2023executing_mld,jiang2023motiongpt}, audio~\citep{yi2023generating_talkshow}, and scene~\citep{zhang2020generating} inputs. %Text, being an accessible and versatile intermediate, holds particular promise for users and aligns with our research focus. 
On the modeling front, contemporary human motion generation primarily relies on two dominant paradigms: auto-regressive methods operating in discrete spaces~\citep{zhang2023generating_t2mgpt} and non-auto-regressive approaches grounded in diffusion models~\citep{chen2023executing_mld,tevet2022human_mdm,zhang2022motiondiffuse}. The former often accumulates errors and demands iterative frame generation, resulting in time-intensive processes. By contrast, the latter diffusion-approach offers stability, efficient training, and seamless integration of guidance signals, but is hindered by slow sampling speeds~\citep{salimans2022progressive}. Quoting Chen \etal~\cite{chen2023executing_mld}: ``a typical diffusion-based method MDM~\citep{tevet2022human_mdm} requires 24.74 seconds for average inference and up to a minute for maximum inference on a single V100.''  Although various acceleration techniques have been explored, they have yet to fundamentally alter the  nature of diffusion models~\citep{ho2020denoising,song2021scorebased_sde}.

Recently, a novel generative model, known as flow matching~\citep{lipman2022flow,RectifiedFlow_ICLR23,albergo2022building,neklyudov2023action}, has garnered significant attention. In contrast to the curvature trajectory of diffusion models~\cite{karras2022elucidating}, this model is particularly effective at preserving straight trajectories during the generation process by an ordinary differential equation (ODE) solver. It achieves this by regressing the linearly interpolated vector field in the training process. This makes it a promising alternative for addressing challenges related to trajectories, which are commonly encountered in diffusion models.  Although flow matching has been explored for modalities as diverse as video~\citep{video_fm}, audio~\citep{le2023voicebox}, and point clouds~\citep{wu2023fast}, its application in the context of motion generation remains unexplored. In this paper, we introduce the flow matching model into the task of human motion generation. Remarkably, we achieve equivalent performance to previous diffusion-based methods, while significantly reducing the required sampling steps from thousand to ten, facilitating more practical use, see~\Cref{fig:teaser}.

Besides the sampling speedup, recent advancements in generative models have introduced techniques for data editing and imputation, enabling the modification and restoration of data while preserving data distributions. A typical approach for image inpainting, referred to as ``replacement'' as highlighted in~\cite{song2021scorebased_sde,ho2022video}, leverages the equivalence of the forward and backward passes within diffusion models to align the sampling process with known data segments and generates the unknown portions correspondingly. Nevertheless, a similar method's exploration within the context of flow matching has remained unexplored. 
In human motion generation, we utilize flow matching's straight trajectory property to align known motion segments with the trajectory, while ensuring consistency in generating unknown motions. We provide a motion prefix and sometimes also a suffix to guide our model under textual conditions for specific motion generation that preserves consistency. Additionally, we perform inpainting in the joint space, enabling semantic editing of body parts without affecting others. 

Hence, our contributions in this paper encompass three key aspects. First, we introduce the Motion Flow Matching Model, which is a straightforward flow matching-based generative model for human motion generation. Our model strikes an optimal balance between generation quality and sampling steps across various tasks, including text-to-motion and action-to-motion generation. To the best of our knowledge, this is the inaugural application of flow matching in human motion generation.  
Second, we propose a simple training-free editing method named ``sampling trajectory rewriting'', facilitating editing capabilities based on flow matching, not previously explored in the literature. These editing techniques are versatile and well-suited for in-between motion editing, upper body manipulation, as well as motion interpolation tasks.  Lastly, our experimental outcomes establish state-of-the-art Fréchet Inception Distance (FID) performance on the KIT dataset with a minimal number of sampling steps. %Additionally, we attained competitive results on the HumanML3D and HumanAct12 (action-to-motion) datasets.

\section{Related Work}
\label{sec:intro}
%\vspace{-5pt}

\subsection{Diffusion and Flow-based Generative Models}

Diffusion models~\citep{sohl2015deep,ho2020denoising,song2021scorebased_sde} have found broad applications in computer vision, spanning image~\citep{rombach2022high_latentdiffusion_ldm}, audio~\citep{liu2023audioldm}, video~\citep{ho2022video,blattmann2023align_videoldm}, and point cloud generation~\citep{luo2021diffusion}. Even though they have presented high fidelity in generation, they do so at the cost of sampling speed, usually demanding thousands of sampling steps.  Hence, several works propose more efficient sampling techniques for diffusion models, including distillation~\citep{salimans2022progressive,song2023consistency}, noise schedule design~\citep{kingma2021variational,nichol2021improved,preechakul2022diffusion_autoencoder}, and training-free sampling~\citep{song2020denoising_ddim,karras2022elucidating,lu2022dpm,liu2022pseudo_pndm}. Nonetheless, it is important to highlight that existing methods have not fully addressed the challenge of curve trajectory modeling within diffusion models from root, as their forward pass is inherently designed to exhibit curvature in SDE, following either a \emph{Variance Preserving} SDE~\citep{ho2020denoising} or a \emph{Variance Exploding} SDE~\citep{nichol2021improved}.%whether Variance-Preserving SDE or Variance-Explosion SDE,

A recent entrant, known as flow matching~\citep{lipman2022flow,RectifiedFlow_ICLR23,albergo2022building,neklyudov2023action}, has gained prominence for its ability to maintain straight trajectories during generation by an ODE solver, positioning it as an apt alternative for addressing trajectory-related issues encountered in diffusion models. The versatility of flow matching has been showcased on modalities including image~\citep{lipman2022flow}, video~\citep{video_fm}, audio ~\citep{le2023voicebox},  point clouds~\citep{wu2023fast}, and Riemannian manifolds~\citep{chen2023riemannian}. This underscores its capacity to address the inherent trajectory challenges associated with diffusion models, aligning naturally with the limitations of slow sampling in the current diffusion-based motion generation solutions. % based on diffusion models.

% Meanwhile, users often require the ability to perform editing, and the exploration of training-free editing techniques has been a significant focus within unconditional diffusion models, such as SDEdit~\citep{meng2021sdedit} and ILVR~\citep{choi2021ilvr}, all of which rely on generative priors. Similarly, in the realm of conditional diffusion models, methods relying on cross-attention~\citep{p2p,nulltextinversion} have been employed. However, it's worth noting that these approaches predominantly center around SDE-based methods.
Meanwhile, the exploration of training-free editing techniques has been a significant focus within unconditional diffusion models, such as SDEdit~\citep{meng2021sdedit} and ILVR~\citep{choi2021ilvr}, all of which rely on generative priors. Similarly, in the realm of conditional diffusion models, methods relying on cross-attention~\citep{p2p,nulltextinversion} have been employed. However, it's worth noting that these approaches predominantly center around SDE-based methods.
Conversely, the exploration of ODE-based editing, particularly within the context of newly proposed flow matching models, remains relatively underexplored. This serves as a compelling motivation for our investigation into human motion synthesis.

\subsection{Human Motion Generation}
%\vspace{-5pt}

Human motion synthesis involves generating diverse and realistic human-like motion. The data can be represented using either keypoint-based~\citep{zhang2021we,zanfir2021thundr,ma20233d} or rotation-based~\citep{loper2023smpl,pavlakos2019expressive} representations. In this paper, we choose the rotation-based representation due to its representation efficiency resulting from the inductive bias of the human kinematic tree.
% Instead of focusing on unconditional motion generation, researchers usually use conditional inputs, such as text~\citep{petrovich2022temos,zhang2022motiondiffuse,tevet2022human_mdm,guo2022tm2t,jiang2023motiongpt}, action~\citep{petrovich2021action,guo2020action2motion,tevet2022human_mdm,chen2023executing_mld}, and incomplete motion~\citep{ma2022multi,tevet2022human_mdm}.
In addition to unconditional motion generation, conditional inputs are used such as text~\citep{petrovich2022temos,zhang2022motiondiffuse,tevet2022human_mdm,guo2022tm2t,jiang2023motiongpt}, action~\citep{petrovich2021action,guo2020action2motion,tevet2022human_mdm,chen2023executing_mld}, and incomplete motion~\citep{ma2022multi,tevet2022human_mdm}.
%
% Among them, text is the most informative and useful medium for users, and it is the focus of our work. MDM~\citep{tevet2022human_mdm} proposes a diffusion-based generative model~\citep{ho2020denoising} separately trained on several motion tasks. MotionGPT~\citep{jiang2023motiongpt} presents a principled approach to the interaction between motion and language. RemoDiffuse~\citep{zhang2023remodiffuse} explores motion generation from a retrieval perspective, drawing inspiration from~\cite{blattmann2022retrieval}. T2M-GPT~\citep{zhang2023generating_t2mgpt} investigates a generative framework based on VQ-VAE and a Generative Pre-trained Transformer (GPT) for motion generation.
In this work, we mainly explore the text condition as the most informative and user-applicable medium. MDM~\citep{tevet2022human_mdm} proposes a diffusion-based generative model~\citep{ho2020denoising} separately trained on several motion tasks. MotionGPT~\citep{jiang2023motiongpt} presents a principled approach to the interaction between motion and language. RemoDiffuse~\citep{zhang2023remodiffuse} explores motion generation from a retrieval perspective, drawing inspiration from~\cite{blattmann2022retrieval}. T2M-GPT~\citep{zhang2023generating_t2mgpt} investigates a generative framework based on VQ-VAE and a Generative Pre-trained Transformer (GPT) for motion generation.
MLD~\citep{chen2023executing_mld} advances the latent diffusion model~\citep{rombach2022high_latentdiffusion_ldm} to generate motions based on different conditional inputs. Our work introduces a novel model into human motion synthesis, with a primary objective of efficient sampling.

% Additionally, the motion completion task generates motion conditioning on partial motions, such as classical motion prediction \citep{zhang2021we, ma2022multi} or motion in-between \citep{tevet2022human_mdm}, which generates the intermediate motion while partial parts are fixed. Prior research efforts have primarily explored this aspect through either SDE-style diffusion models~\citep{tevet2022human_mdm} or classical distribution alignment techniques~\citep{ma2022multi}. In contrast, our paper takes a novel approach 9by examining it from the perspective of the ODE sampling process.
Additionally, the motion completion task generates motions given partial inputs, such as classical motion prediction \citep{zhang2021we, ma2022multi} or motion in-between \citep{tevet2022human_mdm}. Prior research efforts have primarily explored either SDE-style diffusion models~\citep{tevet2022human_mdm} or classical distribution alignment techniques~\citep{ma2022multi}. In contrast, our paper takes a novel approach by examining it from the perspective of the ODE sampling process.

\begin{figure*}
    \centering
    \includegraphics[width=1.0\textwidth]{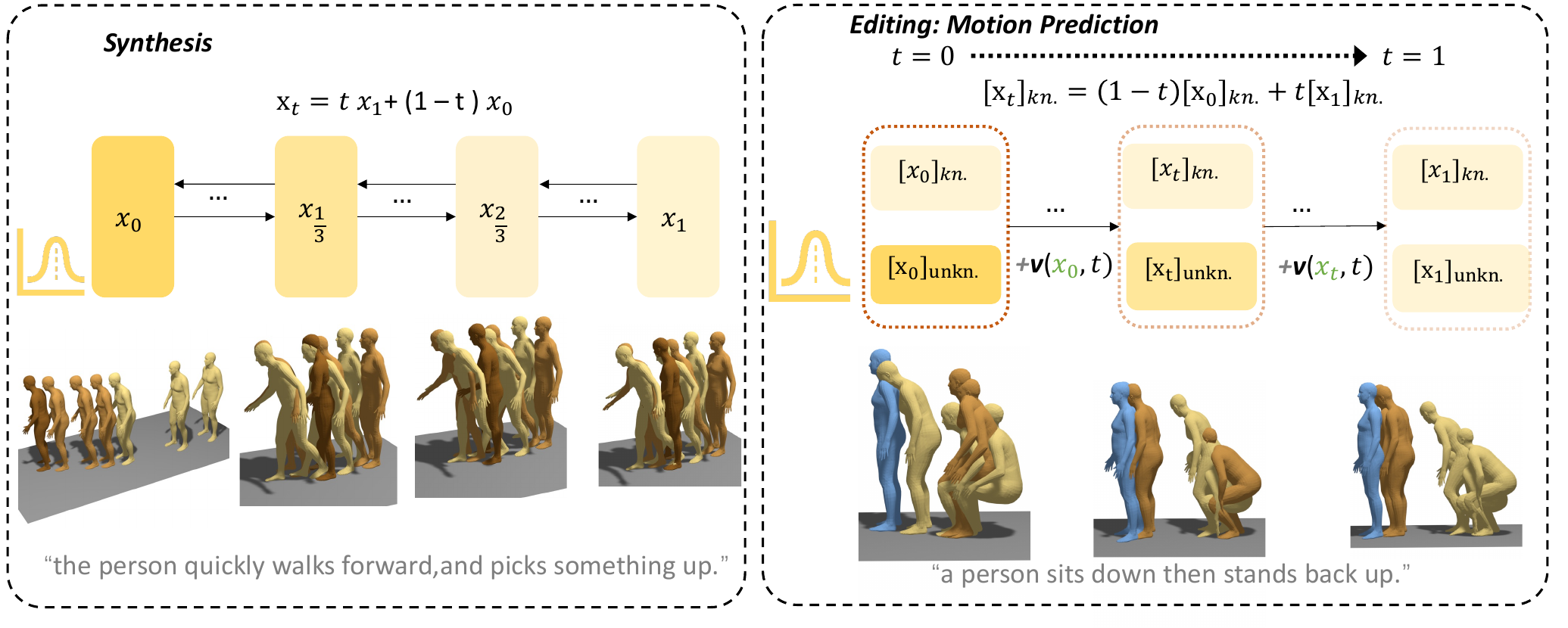}
    \vspace{-20pt}
    \caption{\textbf{Our Motion Flow Matching framework for human motion synthesis and editing.} \emph{Synthesis}: starting with a motion feature $\bx_1$ and a randomly sampled prior Gaussian $\bx_0$, we gradually corrupt the motion feature using simple interpolation: $\bx_t = t \bx_1 + (1 - t) \bx_0$ (omitting $\sigma_{min}$ for simplicity). \emph{Editing}: in editing tasks, we aim to synthesize unknown dimensions in the motion representation while keeping the known components. We sample a Gaussian vector $\bx_0$ and apply sampling trajectory rewriting to the known part of the motion as $[\bx_t]_{\text{known}} = (1-t)[\bx_0]_{\text{known}} + t[\bx_1]_{\text{known}}$ while sampling an ODE solver to adapt the trajectory of the unknown part. As such, the flow $\bv(\bx_t, t)$ is continuously updated with the partially rewritten $\bx_t$, enabling motion editing in various scenarios.
    % \caption{\textbf{Our Motion Flow Matching framework for human motion synthesis and editing.} Starting with a motion feature $\bx_1$ and a randomly sampled prior Gaussian $\bx_0$, we gradually corrupt the motion feature using simple interpolation: $\bx_t = t \bx_1 + (1 - t) \bx_0$ (omitting $\sigma_{min}$ for simplicity). During editing, we sample a Gaussian vector $\bx_0$ and apply sampling trajectory rewriting using the equation $[\bx_t]_{\text{known}} = (1-t)[\bx_0]_{\text{known}} + t[\bx_1]_{\text{known}}$ while sampling an ODE solver. The vector $\bv(\bx_t, t)$ is continuously updated with the newly partially rewritten $\bx_t$. This allows us to achieve motion editing in various scenarios.
    % 
    %The human motion below illustrates the estimation of $\bx_1$ at a specific time step $t$, following Appendix~\Cref{alg:traj_rewrite_python} \tao{change omega}.
    }
    \label{fig:framework}
\end{figure*}
%\vspace{-15pt}

%\vspace{-10pt}
\section{Method}
\label{sec:flow}

In this section, we delve into the framework of motion flow matching, as illustrated in~\Cref{fig:framework}. Then, we present our training-free sampling trajectory rewriting techniques, customized for the flow matching model.

%\vspace{-10pt}
\subsection{Motion Flow Matching}

Our primary goal is to synthesize a human motion $\bx$, with a condition $\bc$ such as a natural language description or an action label. The human motion is formed as a sequence of human poses $\bx = \{\bx^i\}_{i=1}^M$, where the pose in each frame $\bx^i$ is represented by the 3D position or rotation of joints. To achieve this, we build a generative model $f$ parametrized by $\theta$ to synthesize the motion $\bx = f(\bz, \bc; \theta)$, given $\bz$ as a sampled Gaussian noise vector. %, where $\bc$ can encompass diverse notations, such as textual descriptions or action labels.

% \paragraph{Motion representation.}
% In the domain of motion generation, our primary goal is to synthesize human poses, represented as $\bu$, over a temporal continuum. The representation of motion features in each frame can be approached through two fundamental paradigms: a keypoint-based representation~\citep{ma20233d} and a rotation-based representation~\citep{loper2023smpl,pavlakos2019expressive}. While both modalities offer valuable insights, this study gives precedence to the rotation-based representation due to its inherent efficiency, such that $\bu = g(\bx)$, where $\bx$ denotes the rotation-based representation. Consequently, we formulate our core research problem as follows: given a randomly sampled noise vector $\bx_0$, our principal objective is to train a generative model parametrized by $\theta$, which is expressed as $\bx = f(\bx_0, \bc; \theta)$, where $\bc$ can encompass diverse notations, such as textual descriptions or action labels. Following the completion of the training process, the desired motion $\bu$ can be reconstructed using $\bu = g(\bx)$ based on a kinematic tree structure. For a more comprehensive introduction to the representation, please refer to ~\Cref{sup:representation}.

\vspace{-10pt}
\paragraph{Flow matching generation.}
The generative model $f(\cdot)$ can be expressed using either an autoregressive style~\citep{zhang2023generating_t2mgpt} or non-autoregressive models~\citep{tevet2022human_mdm}. However, both approaches face challenges, such as error accumulation~\citep{gong2022future} or a slow sampling speed.  
To alleviate those issues, we have chosen to adopt a new generative model called Flow Matching. Given a set of samples from an unknown data distribution $q(\mathbf{x})$, the goal is to learn a \textit{flow} that pushes the simple prior density $p_0(\mathbf{x}) {=} {\cal N}(\mathbf{x} \,|\, 0, 1)$ towards a more complicated distribution $p_1(\mathbf{x}) {\approx} q(\mathbf{x})$ along the probability path $p_t(\mathbf{x})$. Formally, this is denoted using the push-forward operation as $p_t {=} [\phi_t]_* p_0$.
Following this definition, the motion data $\bx$ is represented as $\bx_{t=1}$ or $\bx_{1}$ while the noise vector $z$ that generates this motion is denoted as $\bx_{t=0}$ or $\bx_0$.
The time-dependent flow can be constructed via a vector field $\mathbf{v}(\mathbf{x},t): \mathbb{R}^d \times  [0, 1]  \rightarrow \mathbb{R}^d$ that defines the flow via the neural ODE: %ordinary differential equation (ODE):
\begin{align}\label{eq:ode}
    \dot \phi_t(\mathbf{x}) = \mathbf{v}(\phi_t(\mathbf{x}), t), \quad \quad
    \phi_0(\mathbf{x}) &= \mathbf{x}_0 .
\end{align}
Given a predefined probability density path $p_t(\mathbf{x})$ and the corresponding vector field $\mathbf{w}_t(\mathbf{x})$, one can parameterize $\mathbf{v}(\mathbf{x}_{t},t)$ with a neural network, parameterized by $\theta$,  and solve
\begin{align}\label{eq:fm}
    \min_{\theta} \mathbb{E}_{t, p_t(\mathbf{x})} \| \mathbf{v}(\mathbf{x}_t,t;\theta) - \mathbf{w}_t(\mathbf{x}) \|^2.
\end{align}
%
%\subsection{Framework}
% Instead of employing generative modeling within the pose coordinates $\bu$, our approach involves its application to motion features $\bx$. These motion features encompass local joint positions, velocities, and rotations in the root space, as well as global translations and rotations.
%
%\cs{It seems this is a general description that should appear after Eq. 2?}
%
%\subsection{Framework}

\paragraph{Framework.} Noticeably, directly optimizing~\Cref{eq:fm} is infeasible, because we do not have access to $\mathbf{w}_t(\mathbf{x})$ in closed form. Instead, ~\citet{lipman2022flow,RectifiedFlow_ICLR23,albergo2022building,neklyudov2023action} propose to use the conditional vector field $\mathbf{w}_t(\mathbf{x} \,|\, \mathbf{x}_1)$ as the target, which corresponds to the conditional flow $p_t(\mathbf{x} \,|\, \mathbf{x}_1)$. Importantly, they show that this new \textit{conditional Flow Matching} objective, defined as: 
\begin{align}
    \min_{\theta} \mathbb{E}_{t, p_t(\mathbf{x} \,|\, \mathbf{x}_1), q(\mathbf{x}_1)} \| \mathbf{v}(\mathbf{x}_t, t;\theta) - \mathbf{w}_t(\mathbf{x} \,|\, \mathbf{x}_1) \|^2,
\end{align}
has the same gradients as \Cref{eq:fm}. 
By defining the conditional probability path as a linear interpolation between $p_0$ and $p_1$, all intermediate distributions are Gaussians of the form $p_t(\mathbf{x} \,|\, \mathbf{x}_1) {=} {\cal N}(\mathbf{x} \,|\, t \bx_1, 1-(1-\sigma_{\text{min}})t)$, where $\sigma_{\text{min}}>0$ is a small amount of noise around the sample $\bx_1$: $\bx_t = t \bx_1 + [1 - (1 - \sigma_\text{min}) t] \bx_0$.
The corresponding target vector field is:
\begin{align}\label{eq:u}
    \mathbf{w}_t(\mathbf{x} \,|\, \mathbf{x}_1) = \mathbf{x}_1 - (1 - \sigma_{\text{min}}) \mathbf{x}.
\end{align}
Learning the straight trajectory improves the training and sampling efficiency compared to diffusion paths. 
When we need extra condition signals $\bc$, we can directly insert them into the vector field estimator $\mathbf{v}(\mathbf{x}_t, t, \bc;\theta)$.
Overall, the framework of flow matching allows it to generate samples by first sampling $\mathbf{x}_0 \sim {\cal N}(\mathbf{x} \,|\, 0, 1)$ and then solving \Cref{eq:ode} using an off-the-shelf numerical ODE solver~\citep{Runge, Kutta, alexander1990solving}. In the end, we can formulate $f$ as 
$\bx=f(\bx_0) = \text{ODESolve}(\bx_0,\bc;\theta)_{t:0\rightarrow1}$.
%$\bx=f(\bx_0); f(\bx_0)= \text{ODESolve}(\bx_0,\bc;\theta)_{t:0\rightarrow1}$.

%For a complete description of flow matching, we direct the reader to Appendix~\Cref{alg:flow_matching_pseudo}. 

%\cs{So? What should the reader conclude at this point after reading all this background on flow matching? What will this paper offer? What is new?}

%Similarly, we can invert a data point to get its corresponding latent noise by solving the ODE in the other direction. \tao{add more stuffs to hint the editting operation.}

%\subsection{Sampling} 

%to introduce the ODE solver sampling and editing tricks.
%\tao{method highlight: 1). only need to edit the top 10\% percentage of the timesteps. 2). can be adapted to different adaptive step-size ODE solvers.}

\paragraph{Sampling.} After the training of the neural velocity field $\bv(\bx_t,t, \bc;\theta)$, the generation of samples is facilitated through practical discretization of the ordinary differential equation (ODE) process outlined in~\Cref{eq:ode} by employing an ODE solver. Using the Euler ODE solver as an illustration, this discretization method entails dividing the process into $N$ steps, resulting in the following expression:
\begin{equation}
\bx_{(\hat{t}+1)/N} \leftarrow \bx_{\hat{t}/N} + \frac{1}{N} ~\bv(\bx_{\hat{t}/N}, \frac{\hat{t}}{N},\bc;\theta),
\label{eq:sample}
\end{equation}
where the integer time step $\hat{t} = 0,1,\cdots,N-1$ such that $t = \hat{t}/N$.
%It is worth noting that a larger value of $N$ results in a more accurate solver, intuitively speaking. 
Additionally, for enhanced efficiency in sampling, alternative approaches such as adaptive step-size ODE solvers~\cite{Runge, Kutta} can be considered, which can significantly reduce the computational time required.

Given that vector field regression in flow matching emulates noise prediction techniques used in diffusion models~\citep{zheng2023improved}, we further investigate the incorporation of classifier-free guidance~\citep{ho2021classifier} in flow matching. This entails introducing random dropout to the conditional signals. In practice, our network learns both the conditioned and unconditioned distributions by randomly setting $\bc=\emptyset$ for $10\%$ of the samples. This configuration effectively causes $\bv(\bx_t, t,\emptyset;\theta)$ to approximate $p(\bx_1)$, signifying that a predominant portion of the network's capacity is dedicated to conditional sampling (90\%) rather than unconditional sampling (10\%).
Subsequently, we conduct the sampling according to the equation:
\begin{equation}
\setlength{\thinmuskip}{0mu}
\setlength{\medmuskip}{0mu}
\setlength{\thickmuskip}{0mu}
{
\bv_s(\bx_t, t, \bc;\theta) = \bv(\bx_t, t, \emptyset;\theta) + s \cdot \left(\bv(\bx_t, t, \bc;\theta) - \bv(\bx_t, t, \emptyset;\theta)\right),
}
\end{equation}
where $s$ indicates the guidance strength that balances the trade-off between diversity and fidelity.

% \subsection{The network dynamics}
%\subsection{Modelling the Flow network dynamics for motion}

%\cs{Some reader guidance is needed here on what to expect here, the text is unguided and unfocused. A figure is introduced without any context and you switch from formal description to informal description, which makes it weak. As if you only apply an existing model to a new problem.}

%When we finish the ODE sampling to recover the motion feature $\bx_1$, we can further convert it to human pose coordinate $\bu_1$ by SMPL model~\cite{loper2023smpl}: $\bu = \texttt{SMPL} ( \bx)$.

\paragraph{The network.}  The dynamics of the flow matching model $\theta$ is founded upon a simple encoder-only architecture based on the Transformer~\citep{vaswani2017attention}.
The Transformer architecture is meticulously designed to possess temporal awareness, facilitating the acquisition of knowledge pertaining to motions of varying durations. Its efficacy within the motion domain has been empirically substantiated~\citep{tevet2022human_mdm,duan2021single,aksan2021spatio}.
For model inputs, $\bx$, the time-step $t$, and the condition code $\bc$, undergo individual fully connected projections into the Transformer dimension via feed-forward networks. These projections are subsequently aggregated to yield the token $\bh_{[x_t,t,c]}$. 
Each frame of the noisy input data $\bx_t$ is subject to linear projection into the Transformer dimension and is summed with a positional embedding. The detailed structure can be found in Appendix~\Cref{sup:impl_detail}. %~\Cref{fig:transformer}.

%The transformer operates on both $\bz_{[x_t,t,c]}$ and the projected motion frame features.

%It is noteworthy that, except for the initial output token corresponding to $\bz_{[x_t,t,c]}$, the encoder's outcomes are re-projected back into the original motion dimensions and serve as predictions of  $\bv({\bx}_{t},t, \bc;\theta)$.
%
%

% CLIP~\citep{radford2021learning} text encoder, 

%\vspace{-18pt}
%\vspace{-10pt}
\subsection{Motion Editing}

\paragraph{Editing tasks.}
We mainly explore the following editing operations: 1) Motion in-between based on prefix and suffix in the temporal domain, 2) Motion prediction based on prior prefix motion in the temporal domain, 3) Motion interpolation with a gap of frames, and 4) Editing body parts in the spatial domain.
The editing operations involve only the sampling process in inference, without any additional training steps. 
In temporal editing (in-between and motion prediction), the input consists of the prefix and suffix frames of the motion sequence.  In the spatial setting, we hope that body parts can be re-synthesized while preserving the rest of the body. Editing can be performed conditionally or unconditionally, with the option to set $\bc=\emptyset$ in the latter case.

%\tao{add random initialized noise or x-inverted noise?}

% \textbf{Motion Editing by sampling trajectory rewriting.} In  diffusion models, a technique known as ``replacement'' is employed during the sampling process to address data imputation challenges~\citep{ho2022video, song2021scorebased_sde}. However, the flow matching model takes a fundamentally different approach compared to diffusion models. In diffusion models, the network gradually corrupts data following either a variance-preserving (VP-SDE)~\citep{ho2020denoising} or variance-explosion (VE-SDE) approach~\citep{song2021scorebased_sde}. In contrast, flow matching initially knows the target distribution sample $\bx_0$ and then ``causalizes'' the intermediate data sample $\bx_t$ through simple interpolation: $\bx_t = t \bx_1 + [1 - (1 - \sigma_{min}) t] \bx_0$. This stands in stark contrast to diffusion models, where $\bx_0$ becomes known only after conducting sufficient \textit{stochastic} forward steps to sample it.

\paragraph{Motion editing by sampling trajectory rewriting.} In diffusion models, a technique known as ``replacement'' is employed during the sampling process to address data imputation challenges~\citep{ho2022video, song2021scorebased_sde}. The network gradually corrupts data following either a variance-preserving (VP-SDE)~\citep{ho2020denoising} or variance-explosion (VE-SDE) approach~\citep{song2021scorebased_sde}. In contrast, flow matching takes a fundamentally different approach such that it initially knows the target distribution sample $\bx_0$ and then ``causalizes'' the intermediate data sample $\bx_t$ through simple interpolation: $\bx_t = t \bx_1 + [1 - (1 - \sigma_{min}) t] \bx_0$. This stands in stark contrast to diffusion models, where $\bx_0$ becomes known only after conducting sufficient \textit{stochastic} forward steps to sample it.

Editing tasks aim to preserve known parts during editing while generating the unknown parts in a manner consistent with the known ones.
Formally, let $\mathcal{M}$ denote a binary boolean mask with the same dimensions as the motion representation $\bx_1$, such that $\mathcal{M}=1$ indicates known corresponding dimension in the given motion $\bx_1$, and $\mathcal{M}=0$ otherwise. 
%
%Let $\Omega(\bx_t)$ and $\bar{\Omega}(\bx_t)$ denote the known and unknown dimensions of $\bx_t$, respectively.
%
Utilizing the principle of pursuing a straight trajectory in flow matching, we consistently enforce the known dimensions as the linear interpolation between $\bx_0$ and $\bx_1$ during sampling steps, while adapting the trajectory of the unknown dimensions from noise. Specifically, in contrast to the standard process for motion synthesis in \Cref{eq:sample}, the sampling process for editing is formalized as:
\vspace{-10pt}
\begin{align}
\label{eq:interp}
\Tilde{\bx}'_{\hat{t}/N} &\leftarrow \underbrace{(1-\mathcal{M})\cdot \Tilde{\bx}_{\hat{t}/N}}_{\vspace{1pt}=[\Tilde{\bx}'_{\hat{t}/N}]_\text{unknown}} + \underbrace{\mathcal{M}\cdot \left( (1- \frac{\hat{t}}{N}) \hspace{0.7pt} \bx_0 + \frac{\hat{t}}{N} ~ \bx_1 \right)}_{=[\Tilde{\bx}'_{\hat{t}/N}]_\text{known}},
%\label{eq:concat}
\\
\label{eq:sampling_edit}
&\Tilde{\bx}_{(\hat{t}+1)/N} \leftarrow \Tilde{\bx}'_{\hat{t}/N} + \frac{1}{N} ~\bv(\Tilde{\bx}'_{\hat{t}/N}, \frac{\hat{t}}{N},\bc;\theta).
\end{align}
\vspace{-0pt}
where $\bx_0$ is a random sampled Gaussian noise that aims to match with the target data $\bx_1$. The intermediate results are denoted with  $\Tilde{\bx}_{\hat{t}/N}$ to discriminate from standard sampling in \Cref{eq:sample},  and $\Tilde{\bx}'_{\hat{t}/N}$ denotes the manipulated $\Tilde{\bx}_{\hat{t}/N}$ during the sampling process.

%At each time step $t$, the vector field can be updated to generate the unknown parts, represented as $\bar{\Omega}(.)$, while preserving the known segments, denoted as $\Omega(.)$. This approach encourages the generation process to maintain consistency with the original input while seamlessly completing the missing components.
%
In our experiments, we found that editing operations do not need to be employed throughout the entire ODE sampling process.  Restricting the sampling trajectory rewriting operation only in early time steps suffices to ensure consistent generation, while granting us more flexibility for editing. Specifically, we set $\varsigma\in [0,1]$ as a threshold such that the rewriting is only applied before that step $t = \frac{\hat{t}}{N} < \varsigma$. Empirically we set $\varsigma = 0.2$ throughout this work.
%For instance, we can perform editing using equations~\ref{eq:interp} and~\ref{eq:concat} within the initial 0.2 time steps employing a fixed time-step ODE solver, while using an adaptive step-size ODE solver for the remaining steps. This approach significantly reduces the sampling time required.
The complete process sampling trajectory rewriting for Euler sampling is shown in \Cref{alg:traj_rewrite}. Under the property of a straight trajectory, the completed unknown part of the motion correctly exhibits the desired marginal distribution by design, and naturally aligns with the known part due to the fully optimized vector field estimator.

%%%%%%%%%%%%%%%%%%%%%%%%%%%%%%%%%%%%%%%

%%%%%%%%%%%%%%%%%%%%%%%%%%%%%%%%%%%%%%%%%%%%%%%%%%%%%%%%%%%%

\begin{algorithm}[t]
\small
\caption{%\small
Euler Sampling algorithm with sampling trajectory rewriting.
}
\label{alg:traj_rewrite}
\begin{algorithmic}[1]
\STATE{\textbf{Input}: $\bx_1$ the original motion (or partial data with valid known dimensions); $\mathcal{M}$ the boolean mask indicating known / unknown dimensions in the motion; $\bv$ and $\theta$ the vector field predictor with pretrained parameters}
\STATE{\textbf{Parameters}: $N$ the number of sampling steps; $\varsigma$ the threshold when sampling trajectory rewriting stops.}
\STATE{Sample $\bx_0 \sim \mathcal{N}(0,1)$ from the Gaussian distribution, $\Tilde{\bx}_0 = \bx_0$ at $\hat{t}=0$.}
\FOR{$\hat{t}=1,2,...,N-1$}
\IF{$\frac{\hat{t}}{N} < \varsigma$}
\STATE{Rewrite $\Tilde{\bx}'_{\hat{t}/N} \leftarrow (1-\mathcal{M})\cdot \Tilde{\bx}_{\hat{t}/N} + \mathcal{M}\cdot \left( (1- \frac{\hat{t}}{N}) \hspace{0.7pt} \bx_0 + \frac{\hat{t}}{N} ~ \bx_1 \right)$}
\STATE{$\Tilde{\bx}_{(\hat{t}+1)/N} \leftarrow \Tilde{\bx}'_{\hat{t}/N} + \frac{1}{N} ~\bv(\Tilde{\bx}'_{\hat{t}/N}, \frac{\hat{t}}{N},\bc;\theta)$.}
\ELSE
\STATE{$\Tilde{\bx}_{(\hat{t}+1)/N} \leftarrow \Tilde{\bx}_{\hat{t}/N} + \frac{1}{N} ~\bv(\Tilde{\bx}_{\hat{t}/N}, \frac{\hat{t}}{N},\bc;\theta)$.}
\ENDIF
\ENDFOR
\STATE{\textbf{Return}: The motion after editing $\Tilde{\bx}_{N/N} = \Tilde{\bx}_{1}$.}
\end{algorithmic}
\end{algorithm}
\vspace{-6pt}

\section{Experiment}

\begin{table*}
    \centering
    % \vspace{-10pt}
   \resizebox{0.8\textwidth}{!}{
    \begin{tabular}{l r c c c c c r }
    \toprule

     Methods & NFE $\downarrow$   &RP Top3$\uparrow$ & FID $\downarrow$ & MM-Dist $\downarrow$ &  Diversity $\rightarrow$ & MModality $\uparrow$ & \#params \\

    \midrule
 {Real motion} &- & \et{0.779}{.006} & \et{0.031}{.004} & \et{2.788}{.012} & \et{11.08}{.097} & - & -  \\

%Our VQ-VAE (Recons.) & \et{0.399}{.005} & \et{0.614}{.005} & \et{0.740}{.006} & \et{0.472}{.011} & \et{2.986}{.027} & \et{10.994}{.120} & -  \\
    \midrule
        %Seq2Seq~\cite{lin2018generating}& & \et{0.241}{.006} & \et{24.86}{.348} & \et{7.960}{.031} & \et{6.744}{.106}  & - & {20.6M} \\

        %Language2Pose~\cite{ahuja2019language2pose} & & \et{0.483}{.005} & \et{6.545}{.072} & \et{5.147}{.030} & \et{9.073}{.100} & - & {20.6M} \\

        %Text2Gesture~\cite{bhattacharya2021text2gestures}   && \et{0.338}{.005} & \et{12.12}{.183} & \et{6.964}{.029} & \et{9.334}{.079} & - & {27M}     \\

        %Hier~\cite{ghosh2021synthesis}   && \et{0.531}{.007} & \et{5.203}{.107} & \et{4.986}{.027} & \et{9.563}{.072} & - & {64.5M} \\

        %MoCoGAN~\cite{tulyakov2018mocogan}  && \et{0.063}{.003} & \et{82.69}{.242} & \et{10.47}{.012} & \et{3.091}{.043} & \et{0.250}{.009} & - \\

        %Dance2Music~\cite{lee2019dancing}  && \et{0.086}{.003} & \et{115.4}{.240} & \et{10.40}{.016} & \et{0.241}{.004} & \et{0.062}{.002} & - \\
        
        TM2T~\cite{guo2022tm2t}  & - & \et{0.587}{.005} & \et{3.599}{.153} & \et{4.591}{.026} & \et{9.473}{.117} & \et{{3.292}}{.081} & {317M}  \\
        
       ~\cite{guo2022generating}  &-& \et{0.681}{.007} & \et{3.022}{.107} & \et{3.488}{.028} & \et{10.72}{.145} & \et{2.052}{.107}  & {181M}\\

        T2M-GPT  &-& \et{0.716}{.006} & \et{0.737}{.049} & \et{3.237}{.027} & \et{11.198}{.086} & \et{2.309}{.055} & {247.6M}\\
        
    \midrule
    
        MDM~\cite{tevet2022human_mdm}   & {1,000} & \et{0.396}{.004} & \et{0.497}{.021} & \et{9.191}{.022} & \et{10.847}{.109} & \et{1.907}{.214} & {23M}   \\

        MotionDiffuse   & {1,000} & \et{{0.739}}{.004} & \et{1.954}{.062} & \et{{2.958}}{.005} & \et{11.100}{.143} & \et{0.730}{.013} & {238M}  \\
        MLD~\cite{chen2023executing_mld}& 50 &\et{0.734}{.007}&\et{0.404}{.027}&\et{3.204}{.027}&\et{10.800}{.117}&\et{2.192}{.071}& 26.9M \\
        
    \midrule
          \textbf{Our MFM}  & {10}& \et{0.414}{.006} & \et{{0.359}}{.034} & \et{9.030}{.043} & \et{{11.310}}{.102} & \et{1.220}{.079} &{17.9M}  \\

           \textbf{Our MFM}  & {50}& \et{0.415}{.006} & \et{{0.193}}{.020} & \et{9.041}{.013} & \et{{11.080}}{.108} & \et{1.490}{.056} &{17.9M}  \\
           % \textbf{Our MFM}  & {100}& \et{0.4139}{.006} & \et{\textbf{0.1886}}{.029} & \et{9.0463}{.043} & \et{\textbf{11.06}}{.100} & \et{1.53}{0.067} &{17.9M} \\
    \bottomrule
    \end{tabular}
    }
    %\footnotesize{$^\S$ reports results using ground-truth motion length.} 
    \vspace{-10pt}
    \caption{\textbf{Comparison with state-of-the-art methods on the KIT-ML}~\citep{plappert2016kit} test set. RP Top3 denotes R-Precision Top3. NFE denotes the number of function evaluations. $\rightarrow$ indicates that closer to real is better. Our method excels in FID and Diversity with minimal parameters.
    %\tao{make MFM as our name}\tao{For a full comparison with all methods, pleae check appendix}
    %We compute standard metrics following~\cite{guo2022generating}. For each metric, we repeat the evaluation 20 times and report the average with 95\% confidence interval. \textcolor{red}{Red} and \textcolor{blue}{Blue} indicate the best and the second best result.
    }
    \label{tab:kit}
\end{table*}

\subsection{Datasets and Experimental details}
%\vspace{-7pt}
\label{sec:data}
% Our experimental evaluations are conducted on three established datasets commonly employed for human motion generation tasks: HumanML3D~\cite{guo2022generating}, KIT Motion-Language (KIT-ML)\cite{plappert2016kit} for text-to-motion generation, and an additional action-to-motion generation dataset, HumanAct12\cite{guo2020action2motion}. We adhere to the evaluation protocols outlined in~\cite{guo2022generating}. For further dataset-specific details, please refer to the supplementary materials in Appendix~\Cref{sup:impl_detail}.

\textbf{Three datasets.} \quad Our experimental evaluations are conducted on three established datasets commonly employed for human motion generation tasks: \textit{HumanML3D}~\citep{guo2022generating}, \textit{KIT Motion-Language} (KIT-ML)~\citep{plappert2016kit} for text-to-motion generation, and an additional action-to-motion generation dataset, \textit{HumanAct12}~\citep{guo2020action2motion}. We adhere to the evaluation protocols outlined in~\cite{guo2022generating}. %For further dataset-specific details, please refer to the supplementary materials in ~\Cref{sup:impl_detail}.
We opt for a motion representation following \cite{guo2022generating} for its effectiveness in encoding the motion kinematics. More details are introduced in Appendix~\Cref{sup:representation}. Similar to ~\cite{guo2022generating}, the dataset KIT-ML and HumanML3D are extracted into motion features with dimensions 251 and 263 respectively, which correspond to local joints position, velocity, and rotations in root space as well as global translation and rotations. These features are computed from 21 and 22 joints of SMPL~\citep{loper2023smpl}.

\textbf{Five evaluation metrics.} \quad Our evaluation metrics encompass five key aspects. 1). We evaluate the general number of function evaluations (NFE), denoting the average network forward number.  2), To assess the parameter efficiency of the models, we investigate the number of parameters they contain. 3), For motion quality assessment, we rely on the Frechet Inception Distance (FID), utilizing a feature extractor \citep{guo2022generating} to measure the distance between feature distributions of generated and real motions. 4), To gauge generation diversity, we employ the Diversity metric, which quantifies motion diversity by calculating variance in features extracted from the motions, along with MultiModality (MModality) for assessing diversity within generated motions under the same text description. 5), In terms of text alignment, we utilize the motion-retrieval precision (R Precision) to evaluate the accuracy of matching texts and motions using Top3 retrieval accuracy, while Multi-modal Distance (MM Dist) measures the distance between motions and texts, all based on feature spaces from \cite{guo2022generating}.  

\textbf{Training details.}  \quad
We use the AdamW~\citep{loshchilov2017decoupled_adamw} optimizer with $[\beta_1, \beta_2] = [0.9, 0.999]$, batch size of 256. We train with a learning rate of 1e-4. we employ a timestep of $N=10$ for Euler ODE sampling. However, for sampling trajectory rewriting, we opt for a slightly larger value of $N=30$, but restrict the editing to the initial $t=0.2$ timesteps, effectively modifying the first 6 timesteps only. More details about our implementation aare provided in the Appendix. Code will be made public.

\begin{table*}
    \centering
    %\vspace{-10pt}
   \resizebox{0.8\textwidth}{!}{
    \begin{tabular}{l r c c c c c r }
    \toprule
    Methods & NFE $\downarrow$   &RP Top3$\uparrow$ & FID $\downarrow$ & MM-Dist $\downarrow$ &  Diversity $\rightarrow$ & MModality $\uparrow$ & \#params \\

    \midrule

        {Real motion} &- & \et{0.797}{.002} & \et{0.002}{.000} & \et{2.974}{.008} & \et{9.503}{.065} & - &   \\
        %Our VQ-VAE \small{(Recons.)} & & \et{0.785}{.002} & \et{0.070}{.001}& \et{3.072}{.009} & \et{9.593}{.079} & -  \\
    \midrule
       % Seq2Seq~\cite{lin2018generating}& & \et{0.396}{.002} & \et{11.75}{.035} & \et{5.529}{.007} & \et{6.223}{.061}  & - & {20.6M} \\

       % Language2Pose~\cite{ahuja2019language2pose}& &\et{0.486}{.002} & \et{11.02}{.046} & \et{5.296}{.008} & \et{7.676}{.058} & - & {20.6M} \\

        %Text2Gesture~\cite{bhattacharya2021text2gestures}  & & \et{0.345}{.002} & \et{5.012}{.030} & \et{6.030}{.008} & \et{6.409}{.071} & - & {27M} \\

        %Hier~\cite{ghosh2021synthesis}&  & \et{0.552}{.004} & \et{6.532}{.024} & \et{5.012}{.018} & \et{8.332}{.042} & -  & {64.5M} \\

        %MoCoGAN~\cite{tulyakov2018mocogan}& & \et{0.106}{.001} & \et{94.41}{.021} & \et{9.643}{.006} & \et{0.462}{.008} & \et{0.019}{.000} & - \\

        %Dance2Music~\cite{lee2019dancing} & & \et{0.097}{.001} & \et{66.98}{.016} & \et{8.116}{.006} & \et{0.725}{.011} & \et{0.043}{.001} & - \\

        TM2T~\cite{guo2022tm2t}& - & \et{0.729}{.002} & \et{1.501}{.017} & \et{3.467}{.011} & \et{8.589}{.076} & \et{2.424}{.093} & {317M}  \\

        ~\cite{guo2022generating} & - & \et{0.736}{.002} & \et{1.087}{.021} & \et{3.347}{.008} & \et{9.175}{.083} & \et{2.219}{.074} & {181M}  \\

        T2M-GPT& - & \et{0.685}{.003} & \et{{0.140}}{.006} & \et{3.730}{.009} & \et{9.844}{.095} & \et{{3.285}}{.070} & {247.6M} \\
    MotionGPT &-&\et{0.778}{0.002}&\et{0.232}{.008}& - &\et{{9.520}}{.071} & \et{2.008}{.084}& {220M} \\
    \midrule
    
        MDM~\cite{tevet2022human_mdm} & {1,000} & \et{0.611}{.007} & \et{0.544}{.044} & \et{5.566}{.027} & \et{9.559}{.086} & \et{2.799}{.072} & {23M}  \\

        MotionDiffuse  & {1,000} & \et{{0.782}}{.001} & \et{0.630}{.001} & \et{{3.113}}{.001} & \et{9.410}{.049} & \et{1.553}{.042} & {238M} \\
        MLD~\cite{chen2023executing_mld}& 50 &\et{0.772}{.002}&\et{0.473}{.013}&\et{3.196}{.000}&\et{9.724}{.082}&\et{2.413}{.079}& 26.9M \\
        
        %etr, etbb
    \midrule
         \textbf{Our MFM} & {10} & \et{0.642}{.003} & \et{0.362}{.006} & \et{5.280}{.009} & \et{9.860}{.095} & \et{2.443}{.070} &{17.9M}  \\
         
    \bottomrule
    \end{tabular}
    }

    %\footnotesize{ $^\S$ reports results using ground-truth motion length.} 
    \vspace{-10pt}
    \caption{\textbf{Comparison with state-of-the-art methods on the HumanML3D}~\citep{guo2022generating} test set. RP Top3 denotes R-Precision Top3. NFE denotes the number of function evaluations. $\rightarrow$ indicates that closer to real is better. Our method effectively balances NFE and generative performance.
    %\tao{motiongpt..}
    %We compute standard metrics following Guo ~\cite{guo2022generating}. For each metric, we repeat the evaluation 20 times and report the average with 95\% confidence interval. \textcolor{red}{Red} and \textcolor{blue}{Blue} indicate the best and the second best result. \tao{todo, fid first.}
    }
    \label{tab:humanml3d}

\end{table*}

% \begin{table}
% \centering
% % \vspace{-6pt}
% \caption{\textbf{Evaluation of sampling trajectory rewriting editing.} We edit our motion by randomly generating ${5,000}$ motions, and compare it with the ground truth.  ADE and FDE are joint distances between generation and ground truth.}
% \vspace{-10pt}
% %\resizebox{\columnwidth}{!}{%
% \resizebox{0.7\textwidth}{!}{
% \begin{tabular}{@{}lccc@{}ccccc@{}cccc}
% \toprule
% &\multicolumn{3}{c}{Prediction}&\multicolumn{2}{c}{ Upper body} &\multicolumn{2}{c}{In-between}  \\
% \cmidrule(lr){2-4} \cmidrule(lr){5-6} \cmidrule(lr){7-8}
% &FID $\downarrow$ &ADE$\downarrow$  & FDE$\downarrow$ &FID$\downarrow$  & ADE$\downarrow$  & FID$\downarrow$  & ADE$\downarrow$ \\

%  \toprule
%  MDM~\cite{tevet2022human_mdm}
% & 7.34&5.90&7.50& 8.40 & 5.40 & 3.43 & 4.73\\ 

% %MLD~\cite{chen2023executing_mld}&0.473&0.426&0.432&\textcolor{black}{0.568}\\ 
% %\hline
% \textbf{Our MFM}
% & \textbf{5.79}&\textbf{4.99}&\textbf{5.50}& \textbf{6.46} & \textbf{4.12} &\textbf{2.59} & \textbf{3.32}\\ 
% % \midrule
% % \multirow{2}{*}{MLD-NoVAE} &FID$\downarrow$& & & & 0.544\\ 
% % &AITS$\downarrow$&0.422&0.844&1.665& 8.799\\
% % \midrule

% \bottomrule
% \end{tabular}%
% %}
% }
% % \vspace{-6pt}
% \label{tab:editing}
% % \vspace{-10pt}
% \end{table}

\begin{table}
\centering
% \vspace{-6pt}
%\vspace{-10pt}
\resizebox{0.45\textwidth}{!}{
\begin{tabular}{@{}l@{\hspace{8pt}}c@{\hspace{8pt}}c@{\hspace{8pt}}c@{\hspace{8pt}}c@{\hspace{8pt}}c@{\hspace{8pt}}c@{\hspace{8pt}}c@{\hspace{8pt}}c}
\toprule
&\multicolumn{3}{c}{Prediction}&\multicolumn{2}{c}{Upper body} &\multicolumn{2}{c}{In-between}  \\
\cmidrule(lr){2-4} \cmidrule(lr){5-6} \cmidrule(lr){7-8}
&FID $\downarrow$ &ADE$\downarrow$  & FDE$\downarrow$ &FID$\downarrow$  & ADE$\downarrow$  & FID$\downarrow$  & ADE$\downarrow$ \\

 \toprule
 MDM~\cite{tevet2022human_mdm}
& 7.34&5.90&7.50& 8.40 & 5.40 & 3.43 & 4.73\\ 

%MLD~\cite{chen2023executing_mld}&0.473&0.426&0.432&\textcolor{black}{0.568}\\ 
%\hline
\textbf{Our MFM}
& \textbf{5.79}&\textbf{4.99}&\textbf{5.50}& \textbf{6.46} & \textbf{4.12} &\textbf{2.59} & \textbf{3.32}\\ 
% \midrule
% \multirow{2}{*}{MLD-NoVAE} &FID$\downarrow$& & & & 0.544\\ 
% &AITS$\downarrow$&0.422&0.844&1.665& 8.799\\
% \midrule

\bottomrule
\end{tabular}
}
% \vspace{-6pt}
\label{tab:editing}
\caption{\textbf{Evaluation of sampling trajectory rewriting editing.} We edit our motion by randomly generating ${5,000}$ motions, and compare it with the ground truth.  ADE and FDE are joint distances between generation and ground truth. Our editing excels in all three metrics.}
\vspace{-10pt}
\end{table}

\begin{figure*}
    \centering
    \includegraphics[width=1.0\textwidth]{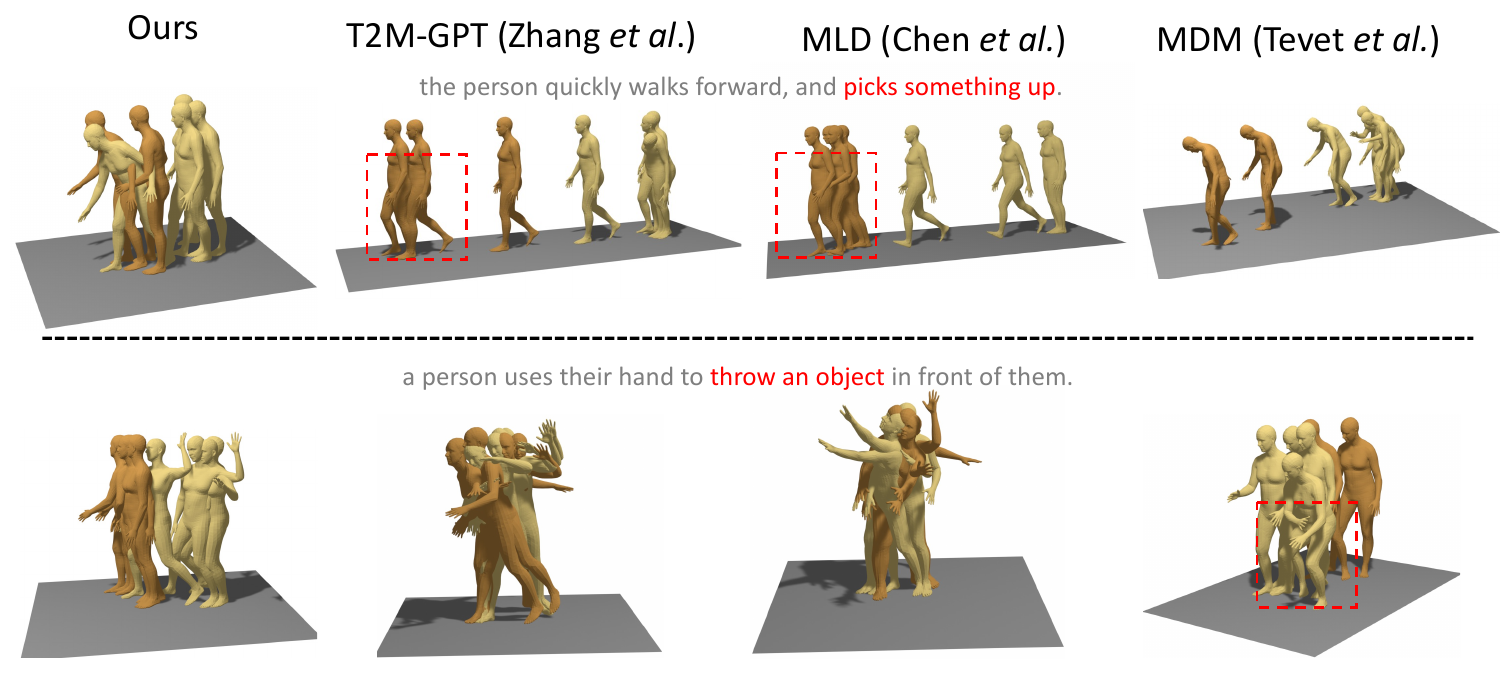}
    \vspace{-24pt}
    \caption{\textbf{Qualitative comparison with baselines.} The generated motion is represented by the bronze frames. Please take note of the  \textcolor{red}{dotted red rectangle}; occasionally, our baseline may struggle to discern the guidance from the prompt, resulting in less fluid motion.
    }
    \label{fig:compare_baseline}
\end{figure*}
\vspace{-6pt}

%\vspace{-10pt}
\subsection{Main Result}
%\vspace{-6pt}
%\tao{To discuss, should move ~\ref{fig:inference_time_humanml} to the very beginning?}

\begin{figure}
    \centering
    \includegraphics[width=0.49\textwidth]{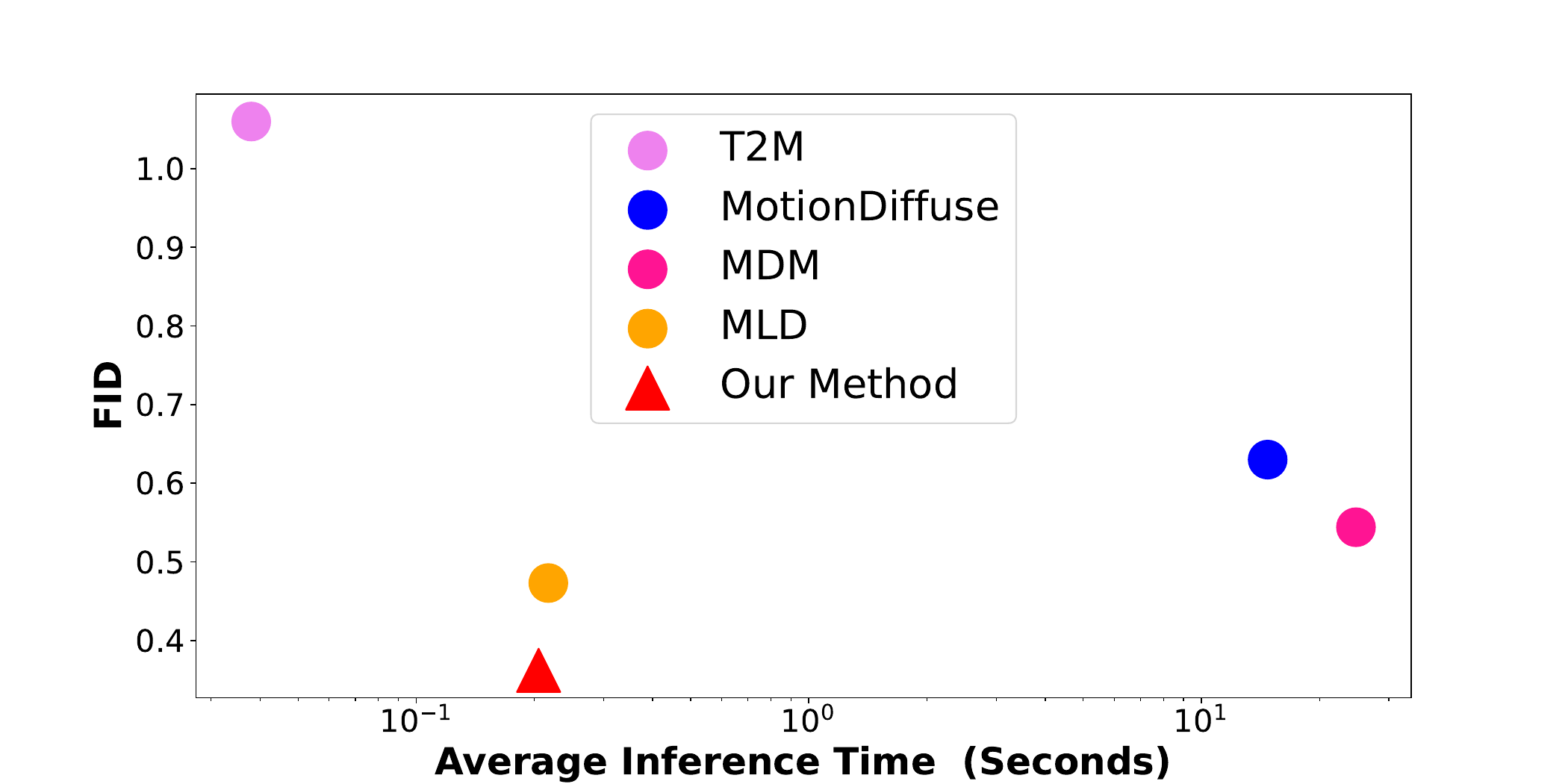}
    \vspace{-10pt}
    \caption{\textbf{Average inference time  comparison with baselines.} Following~\cite{chen2023executing_mld}, we calculate the time on the test set of HumanML3D. All tests are performed on the V100 GPU. The closer the model to  the origin the better.
    }
    \label{fig:inference_time_humanml}
    \vspace{-10pt}
\end{figure}

\textbf{Text-to-Motion.} \quad In the text-to-motion generation, we present our results on the KIT dataset in~\Cref{tab:kit} and on the HumanML3D dataset in~\Cref{tab:humanml3d}. Our approach attains state-of-the-art FID performance on the KIT dataset while requiring minimal function evaluations and a modest number of parameters. 
In the HumanML3D dataset, our results achieve the best FID among all diffusion-based models. Furthermore, we demonstrate a superior balance between inference time and generative performance in HumanML3D, as shown in Figure~\ref{fig:inference_time_humanml}.
These tables clearly highlight the successful achievement of a favorable balance between sampling steps (NFE) and generation performance by our method. 
Notably, GPT-based methods such as T2M-GPT~\citep{zhang2023generating_t2mgpt} and MotionGPT~\citep{jiang2023motiongpt}, which rely on token prediction, tend to require a higher number of network forward evaluations (NFEs), equivalent to the number of tokens, compared to our approach, which uses 10 NFEs. This suggests that our method may be more computationally efficient in terms of NFEs required for motion generation.

In~\Cref{fig:compare_baseline}, we offer a qualitative comparison with three baseline methods: MDM~\citep{tevet2022human_mdm}, MLD~\citep{chen2023executing_mld}, and T2M-GPT~\citep{zhang2023generating_t2mgpt}, where we use the same guidance strength $s{=}2.5$. Our results exhibit enhanced capabilities in capturing the nuanced motion dynamics derived from the input prompts as compared to these baseline approaches. For additional qualitative results in text-to-motion synthesis we refer to Appendix~\Cref{sec:more_results}. %~\Cref{fig:text2motion}.

\textbf{Sampling steps.} \quad In Figure~\ref{fig:teaser}, We investigate the relationship between the number of sampling steps and FID (Fréchet Inception Distance) using the following baselines on the KIT-ML dataset.  1). MDM~\citep{tevet2022human_mdm} with DDPM sampling. 2). MDM with DDIM sampling~\citep{song2020denoising_ddim}. 3). MLD~\cite{chen2023executing_mld} with DDIM sampling. MDM fails to achieve reasonable FID due to the design. Our method converges to a lower FID at the same sampling steps. Our approach is also significantly faster than MLD~\citep{chen2023executing_mld} and MDM~\citep{tevet2022human_mdm}, being approximately 5 times faster than MLD and 100 times faster than MDM. It is worth noting that our Transformer-based architecture can be further optimized using FlashAttention~\citep{dao2022flashattention}.

\textbf{Inference time.} \quad 
%MDM~\cite{tevet2022human_mdm} requires 24.74 seconds for average inference and up to a minute for maximum inferenceon a single V100.
We compared our method with several baselines on the HumanML3D test set in Figure~\ref{fig:inference_time_humanml}. Our approach strikes a superior trade-off between FID and inference time, averaging just 0.11 seconds for a single motion.

\textbf{Action-to-Motion.} \quad We further showcase our action-to-motion generation results in Appendix~\Cref{supp:extra_quantitative}. Our method consistently achieves on-par results with baselines, while requiring significantly fewer function evaluations and better parameter-efficiency, underscoring the efficacy of our approach.

%\vspace{-10pt}
\subsection{Editing by Sampling Trajectory Rewriting}

\begin{figure*}
    \centering    \includegraphics[width=0.8\textwidth]{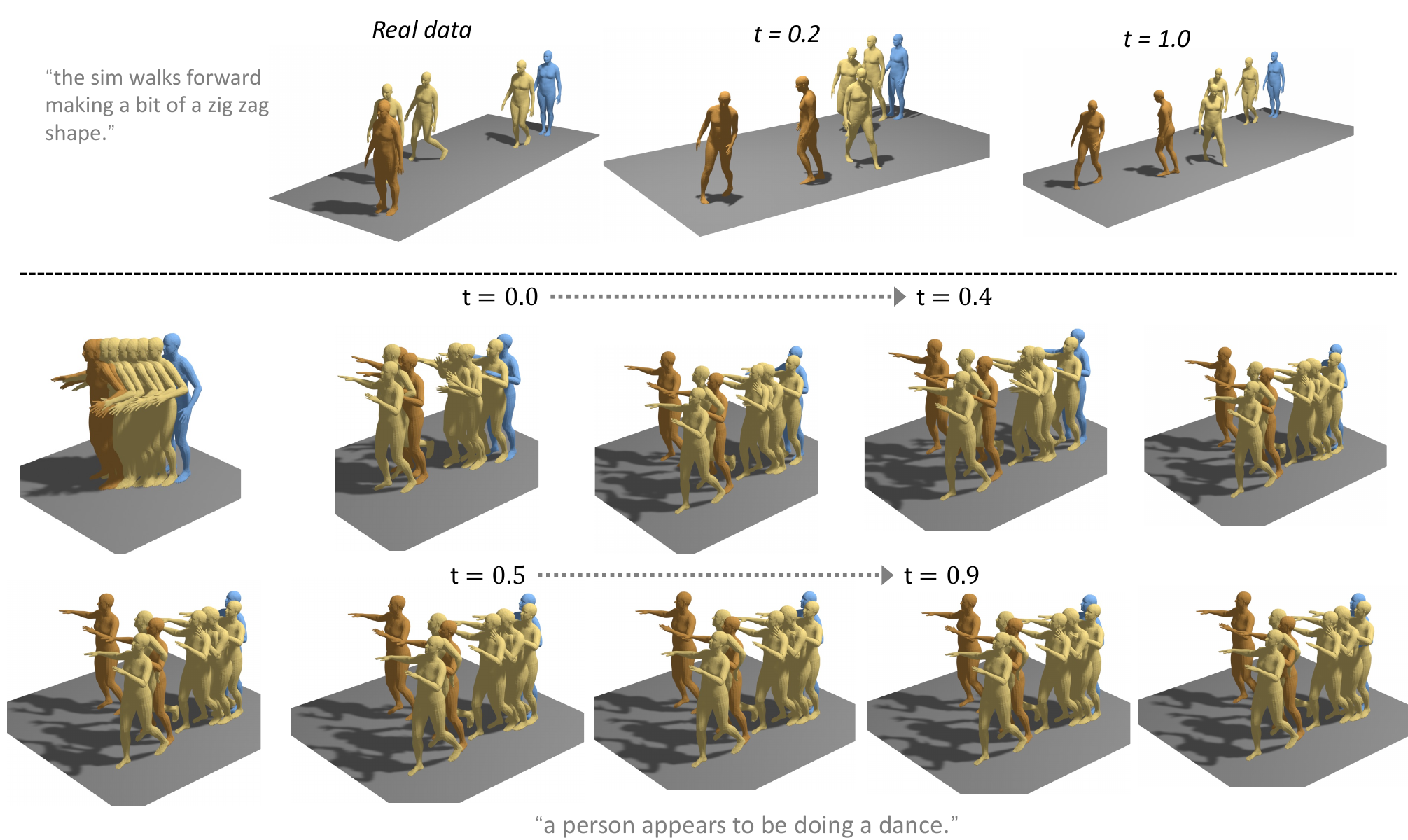}
    \vspace{-10pt}
    \caption{\textbf{Above is a comparison of editing times for the motion prediction task.} Rewriting the trajectory up to $t=0.2$ achieves nearly identical performance compared to rewriting up to $t=1.0$.
    %For a comprehensive overview of editing across various time intervals, please refer to Appendix~\Cref{fig:trajectory_x1_est_10steps}.
    %
    \textbf{Below is the motion prediction: estimation of $\bx_1$ during sampling trajectory rewriting from $t=0.0$ to $t=1.0$} It's worth noting that from the very first time steps, the model can already produce reasonably accurate motion. Furthermore, as time progresses, the estimation of $\bx_1$ gradually aligns more closely with the provided prompt. Remarkably, by the time we reach time step $t=0.2$, the generated human motion exhibits a high level of alignment with the prompt. 
    Light blue frames denote motion input, while bronze frames signify generated motion. The gradient of colors, ranging from light to dark, signifies the passage of time.
    }
    \label{fig:edit_time}
\end{figure*}

%\vspace{-7pt}
\textbf{Qualitative result.} \quad In~\Cref{fig:edit_time},  we demonstrate that editing operations can effectively take place within the previous 0.2 time steps other than the full 1.0 time steps. Furthermore, we provide a demonstration of $\bx_1$ estimation at intervals of 0.1 time steps. Remarkably, we observe that even the initial $\bx_t$ can yield reasonably accurate estimations of $\bx_1$. As time progresses, these estimations gradually align more closely with the provided prompt, eventually stabilizing around $t{=}0.2$. This phenomenon underscores the straight trajectory characteristics of our models.

\begin{figure*}
    \centering
    \includegraphics[width=0.84\textwidth]{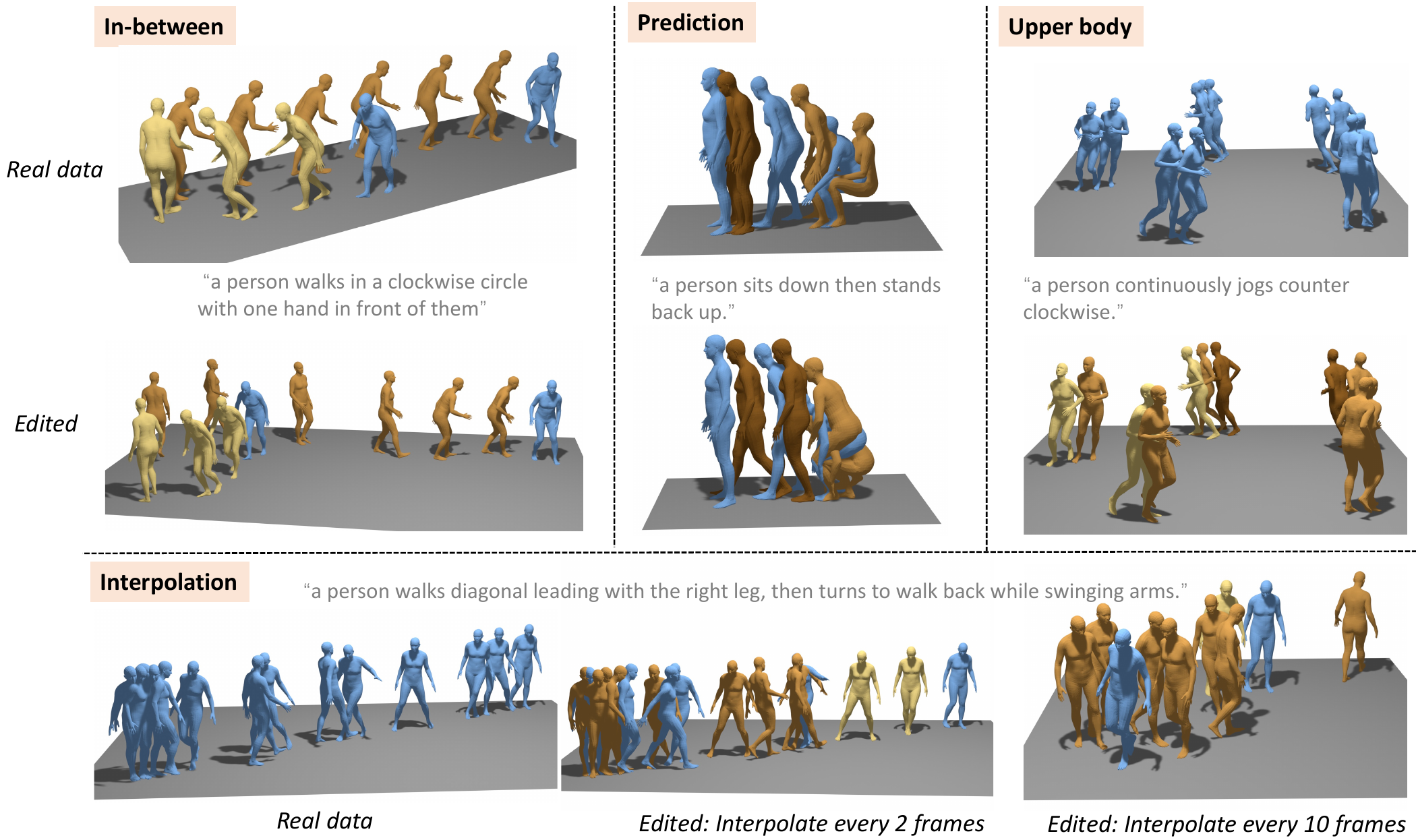}
    \vspace{-8pt}
    \caption{\textbf{Motion Editing by sampling trajectory rewriting.} We focus on showcasing four key editing scenarios in human motion generation: 1) In-between editing, 2) Motion prediction from a partial sequence, 3) Upper body editing while keeping lower body joints fixed, and 4) Interpolating missing motion frames using specified motions. 
    Additionally, we conduct rotations of the views to obtain a more comprehensive global perspective.
    %
   %Light blue frames denote motion input, while bronze frames signify generated motion. The gradient of colors, ranging from light to dark, signifies the passage of time.
    %Motion in-betweening (left+center) can be performed conditioned on text or without any condition using the same model. 
    %On the right, specific body part editing using text is demonstrated: the lower body joints remain fixed to the input motion, while the upper body is modified to match the input text prompt.
    }
    \label{fig:edit_vis_all}
\end{figure*}

Moreover, we investigate several editing operations including in-between, future prediction, upper body, and interpolation editions, as illustrated in~\Cref{fig:edit_vis_all}. We examine the alteration of the text prompt while retaining the known part, serving as a test to evaluate whether our generative models can consistently produce motion that aligns with the preserved known segment.

%Secondly, we delve into motion frame interpolation, with a focus on two scenarios: interpolating every 2 frames (representing a 50\% unknown ratio) and every 10 frames (indicating a 90\% unknown ratio).  We can find our model can dynamically adapt to the provided prompts.

%\vspace{-10pt}
\textbf{Quantitative result.} \quad In our comparison with the baseline method MDM~\citep{tevet2022human_mdm} as illustrated in~\Cref{tab:editing}, we evaluate our sampling trajectory rewriting approach through identical editing operations, revealing its slight performance superiority across FID, Average Displacement Error (ADE), and Final Displacement Error (FDE) metrics, commonly used in motion prediction studies~\citep{zhang2021we}. More details can be found in Appendix~\Cref{sup:impl_detail}.

\section{Conclusion}
%\vspace{-7pt}

In this work, we have introduced a straightforward yet highly effective generative model called Flow Matching to the realm of human motion synthesis. Our results demonstrate a remarkable balance between generation fidelity and sampling steps. Leveraging the inherent property of straight trajectories, we have devised a simple sampling trajectory rewriting technique for training-free editing. In our future endeavors, we intend to extend this sampling trajectory rewriting technique to other domains, such as image editing.

\pagebreak 

% We thank Aozhu Chen to provide several visualization help.
{
    \small
    \bibliographystyle{ieeenat_fullname}
    \bibliography{main}

\begin{thebibliography}{69}
\providecommand{\natexlab}[1]{#1}
\providecommand{\url}[1]{\texttt{#1}}
\expandafter\ifx\csname urlstyle\endcsname\relax
  \providecommand{\doi}[1]{doi: #1}\else
  \providecommand{\doi}{doi: \begingroup \urlstyle{rm}\Url}\fi

\bibitem[Aksan et~al.(2021)Aksan, Kaufmann, Cao, and Hilliges]{aksan2021spatio}
Emre Aksan, Manuel Kaufmann, Peng Cao, and Otmar Hilliges.
\newblock A spatio-temporal transformer for 3d human motion prediction.
\newblock In \emph{3DV}, 2021.

\bibitem[Albergo and Vanden-Eijnden(2023)]{albergo2022building}
Michael~S Albergo and Eric Vanden-Eijnden.
\newblock Building normalizing flows with stochastic interpolants.
\newblock In \emph{ICLR}, 2023.

\bibitem[Alexander(1990)]{alexander1990solving}
Roger Alexander.
\newblock Solving ordinary differential equations i: Nonstiff problems (e. hairer, sp norsett, and g. wanner).
\newblock \emph{Siam Review}, 1990.

\bibitem[Aram~Davtyan and Favaro(2023)]{video_fm}
Sepehr~Sameni Aram~Davtyan and Paolo Favaro.
\newblock Efficient video prediction via sparsely conditioned flow matching.
\newblock In \emph{ICCV}, 2023.

\bibitem[Blattmann et~al.(2022)Blattmann, Rombach, Oktay, and Ommer]{blattmann2022retrieval}
Andreas Blattmann, Robin Rombach, Kaan Oktay, and Bj{\"o}rn Ommer.
\newblock Retrieval-augmented diffusion models.
\newblock In \emph{NeurIPS}, 2022.

\bibitem[Blattmann et~al.(2023)Blattmann, Rombach, Ling, Dockhorn, Kim, Fidler, and Kreis]{blattmann2023align_videoldm}
Andreas Blattmann, Robin Rombach, Huan Ling, Tim Dockhorn, Seung~Wook Kim, Sanja Fidler, and Karsten Kreis.
\newblock Align your latents: High-resolution video synthesis with latent diffusion models.
\newblock In \emph{CVPR}, 2023.

\bibitem[Cervantes et~al.(2022)Cervantes, Sekikawa, Sato, and Shinoda]{cervantes2022implicit}
Pablo Cervantes, Yusuke Sekikawa, Ikuro Sato, and Koichi Shinoda.
\newblock Implicit neural representations for variable length human motion generation.
\newblock In \emph{ECCV}, 2022.

\bibitem[Chen and Lipman(2023)]{chen2023riemannian}
Ricky~TQ Chen and Yaron Lipman.
\newblock Riemannian flow matching on general geometries.
\newblock \emph{arXiv}, 2023.

\bibitem[Chen et~al.(2023)Chen, Jiang, Liu, Huang, Fu, Chen, and Yu]{chen2023executing_mld}
Xin Chen, Biao Jiang, Wen Liu, Zilong Huang, Bin Fu, Tao Chen, and Gang Yu.
\newblock Executing your commands via motion diffusion in latent space.
\newblock In \emph{CVPR}, 2023.

\bibitem[Choi et~al.(2021)Choi, Kim, Jeong, Gwon, and Yoon]{choi2021ilvr}
Jooyoung Choi, Sungwon Kim, Yonghyun Jeong, Youngjune Gwon, and Sungroh Yoon.
\newblock Ilvr: Conditioning method for denoising diffusion probabilistic models.
\newblock In \emph{ICCV}, 2021.

\bibitem[Dao et~al.(2022)Dao, Fu, Ermon, Rudra, and R{\'e}]{dao2022flashattention}
Tri Dao, Dan Fu, Stefano Ermon, Atri Rudra, and Christopher R{\'e}.
\newblock Flashattention: Fast and memory-efficient exact attention with io-awareness.
\newblock In \emph{NeurIPS}, 2022.

\bibitem[Duan et~al.(2021)Duan, Shi, Zou, Lin, Qian, Zhang, and Yuan]{duan2021single}
Yinglin Duan, Tianyang Shi, Zhengxia Zou, Yenan Lin, Zhehui Qian, Bohan Zhang, and Yi Yuan.
\newblock Single-shot motion completion with transformer.
\newblock \emph{arXiv}, 2021.

\bibitem[Gong et~al.(2022)Gong, Lee, Kim, Ha, and Cho]{gong2022future}
Dayoung Gong, Joonseok Lee, Manjin Kim, Seong~Jong Ha, and Minsu Cho.
\newblock Future transformer for long-term action anticipation.
\newblock In \emph{CVPR}, 2022.

\bibitem[Guo et~al.(2020)Guo, Zuo, Wang, Zou, Sun, Deng, Gong, and Cheng]{guo2020action2motion}
Chuan Guo, Xinxin Zuo, Sen Wang, Shihao Zou, Qingyao Sun, Annan Deng, Minglun Gong, and Li Cheng.
\newblock Action2motion: Conditioned generation of 3d human motions.
\newblock In \emph{ACM-MM}, 2020.

\bibitem[Guo et~al.(2022{\natexlab{a}})Guo, Zou, Zuo, Wang, Ji, Li, and Cheng]{guo2022generating}
Chuan Guo, Shihao Zou, Xinxin Zuo, Sen Wang, Wei Ji, Xingyu Li, and Li Cheng.
\newblock Generating diverse and natural 3d human motions from text.
\newblock In \emph{CVPR}, 2022{\natexlab{a}}.

\bibitem[Guo et~al.(2022{\natexlab{b}})Guo, Zuo, Wang, and Cheng]{guo2022tm2t}
Chuan Guo, Xinxin Zuo, Sen Wang, and Li Cheng.
\newblock Tm2t: Stochastic and tokenized modeling for the reciprocal generation of 3d human motions and texts.
\newblock In \emph{ECCV}, 2022{\natexlab{b}}.

\bibitem[Hertz et~al.(2022)Hertz, Mokady, Tenenbaum, Aberman, Pritch, and Cohen-Or]{p2p}
Amir Hertz, Ron Mokady, Jay Tenenbaum, Kfir Aberman, Yael Pritch, and Daniel Cohen-Or.
\newblock Prompt-to-prompt image editing with cross attention control.
\newblock \emph{arXiv}, 2022.

\bibitem[Ho and Salimans(2021)]{ho2021classifier}
Jonathan Ho and Tim Salimans.
\newblock Classifier-free diffusion guidance.
\newblock In \emph{NeurIPS Workshop}, 2021.

\bibitem[Ho et~al.(2020)Ho, Jain, and Abbeel]{ho2020denoising}
Jonathan Ho, Ajay Jain, and Pieter Abbeel.
\newblock Denoising diffusion probabilistic models.
\newblock In \emph{NeurIPS}, 2020.

\bibitem[Ho et~al.(2022)Ho, Salimans, Gritsenko, Chan, Norouzi, and Fleet]{ho2022video}
Jonathan Ho, Tim Salimans, Alexey Gritsenko, William Chan, Mohammad Norouzi, and David~J Fleet.
\newblock Video diffusion models.
\newblock In \emph{arXiv}, 2022.

\bibitem[Jiang et~al.(2023)Jiang, Chen, Liu, Yu, Yu, and Chen]{jiang2023motiongpt}
Biao Jiang, Xin Chen, Wen Liu, Jingyi Yu, Gang Yu, and Tao Chen.
\newblock Motiongpt: Human motion as a foreign language.
\newblock In \emph{NeurIPS}, 2023.

\bibitem[Karras et~al.(2022)Karras, Aittala, Aila, and Laine]{karras2022elucidating}
Tero Karras, Miika Aittala, Timo Aila, and Samuli Laine.
\newblock Elucidating the design space of diffusion-based generative models.
\newblock In \emph{NeurIPS}, 2022.

\bibitem[Kingma et~al.(2021)Kingma, Salimans, Poole, and Ho]{kingma2021variational}
Diederik Kingma, Tim Salimans, Ben Poole, and Jonathan Ho.
\newblock Variational diffusion models.
\newblock In \emph{NeurIPS}, 2021.

\bibitem[Kutta(1901)]{Kutta}
W. Kutta.
\newblock Beitrag zur n\"aherungsweisen {I}ntegration totaler {D}ifferentialgleichungen.
\newblock \emph{Zeit. Math. Phys.}, 1901.

\bibitem[Le et~al.(2023)Le, Vyas, Shi, Karrer, Sari, Moritz, Williamson, Manohar, Adi, Mahadeokar, et~al.]{le2023voicebox}
Matthew Le, Apoorv Vyas, Bowen Shi, Brian Karrer, Leda Sari, Rashel Moritz, Mary Williamson, Vimal Manohar, Yossi Adi, Jay Mahadeokar, et~al.
\newblock Voicebox: Text-guided multilingual universal speech generation at scale.
\newblock In \emph{arXiv}, 2023.

\bibitem[Lipman et~al.(2023)Lipman, Chen, Ben-Hamu, Nickel, and Le]{lipman2022flow}
Yaron Lipman, Ricky~TQ Chen, Heli Ben-Hamu, Maximilian Nickel, and Matt Le.
\newblock Flow matching for generative modeling.
\newblock In \emph{ICLR}, 2023.

\bibitem[Liu et~al.(2023{\natexlab{a}})Liu, Chen, Yuan, Mei, Liu, Mandic, Wang, and Plumbley]{liu2023audioldm}
Haohe Liu, Zehua Chen, Yi Yuan, Xinhao Mei, Xubo Liu, Danilo Mandic, Wenwu Wang, and Mark~D Plumbley.
\newblock Audioldm: Text-to-audio generation with latent diffusion models.
\newblock In \emph{ICML}, 2023{\natexlab{a}}.

\bibitem[Liu et~al.(2022)Liu, Ren, Lin, and Zhao]{liu2022pseudo_pndm}
Luping Liu, Yi Ren, Zhijie Lin, and Zhou Zhao.
\newblock Pseudo numerical methods for diffusion models on manifolds.
\newblock In \emph{ICLR}, 2022.

\bibitem[Liu et~al.(2023{\natexlab{b}})Liu, Gong, and Liu]{RectifiedFlow_ICLR23}
Xingchao Liu, Chengyue Gong, and Qiang Liu.
\newblock Flow straight and fast: Learning to generate and transfer data with rectified flow.
\newblock In \emph{ICLR}, 2023{\natexlab{b}}.

\bibitem[Loper et~al.(2023)Loper, Mahmood, Romero, Pons-Moll, and Black]{loper2023smpl}
Matthew Loper, Naureen Mahmood, Javier Romero, Gerard Pons-Moll, and Michael~J Black.
\newblock Smpl: A skinned multi-person linear model.
\newblock In \emph{Seminal Graphics Papers: Pushing the Boundaries, Volume 2}. 2023.

\bibitem[Loshchilov and Hutter(2019)]{loshchilov2017decoupled_adamw}
Ilya Loshchilov and Frank Hutter.
\newblock Decoupled weight decay regularization.
\newblock In \emph{ICLR}, 2019.

\bibitem[Lu et~al.(2022)Lu, Zhou, Bao, Chen, Li, and Zhu]{lu2022dpm}
Cheng Lu, Yuhao Zhou, Fan Bao, Jianfei Chen, Chongxuan Li, and Jun Zhu.
\newblock Dpm-solver: A fast ode solver for diffusion probabilistic model sampling in around 10 steps.
\newblock \emph{NeurIPS}, 2022.

\bibitem[Luo and Hu(2021)]{luo2021diffusion}
Shitong Luo and Wei Hu.
\newblock Diffusion probabilistic models for 3d point cloud generation.
\newblock In \emph{CVPR}, 2021.

\bibitem[Ma et~al.(2022)Ma, Li, Hosseini, Tomizuka, and Choi]{ma2022multi}
Hengbo Ma, Jiachen Li, Ramtin Hosseini, Masayoshi Tomizuka, and Chiho Choi.
\newblock Multi-objective diverse human motion prediction with knowledge distillation.
\newblock In \emph{CVPR}, 2022.

\bibitem[Ma et~al.(2023)Ma, Su, Wang, Zhu, and Wang]{ma20233d}
Xiaoxuan Ma, Jiajun Su, Chunyu Wang, Wentao Zhu, and Yizhou Wang.
\newblock 3d human mesh estimation from virtual markers.
\newblock In \emph{CVPR}, 2023.

\bibitem[Mahmood et~al.(2019)Mahmood, Ghorbani, Troje, Pons-Moll, and Black]{mahmood2019amass}
Naureen Mahmood, Nima Ghorbani, Nikolaus~F Troje, Gerard Pons-Moll, and Michael~J Black.
\newblock Amass: Archive of motion capture as surface shapes.
\newblock In \emph{ICCV}, 2019.

\bibitem[Mandery et~al.(2015)Mandery, Terlemez, Do, Vahrenkamp, and Asfour]{mandery2015kit}
Christian Mandery, {\"O}mer Terlemez, Martin Do, Nikolaus Vahrenkamp, and Tamim Asfour.
\newblock The kit whole-body human motion database.
\newblock In \emph{ICAR}, 2015.

\bibitem[Meng et~al.(2021)Meng, He, Song, Song, Wu, Zhu, and Ermon]{meng2021sdedit}
Chenlin Meng, Yutong He, Yang Song, Jiaming Song, Jiajun Wu, Jun-Yan Zhu, and Stefano Ermon.
\newblock Sdedit: Guided image synthesis and editing with stochastic differential equations.
\newblock In \emph{ICLR}, 2021.

\bibitem[Mokady et~al.(2023)Mokady, Hertz, Aberman, Pritch, and Cohen-Or]{nulltextinversion}
Ron Mokady, Amir Hertz, Kfir Aberman, Yael Pritch, and Daniel Cohen-Or.
\newblock Null-text inversion for editing real images using guided diffusion models.
\newblock In \emph{CVPR}, 2023.

\bibitem[Neklyudov et~al.(2023)Neklyudov, Brekelmans, Severo, and Makhzani]{neklyudov2023action}
Kirill Neklyudov, Rob Brekelmans, Daniel Severo, and Alireza Makhzani.
\newblock Action matching: Learning stochastic dynamics from samples.
\newblock In \emph{ICML}, 2023.

\bibitem[Nichol and Dhariwal(2021)]{nichol2021improved}
Alexander~Quinn Nichol and Prafulla Dhariwal.
\newblock Improved denoising diffusion probabilistic models.
\newblock In \emph{ICML}, 2021.

\bibitem[Pavlakos et~al.(2019)Pavlakos, Choutas, Ghorbani, Bolkart, Osman, Tzionas, and Black]{pavlakos2019expressive}
Georgios Pavlakos, Vasileios Choutas, Nima Ghorbani, Timo Bolkart, Ahmed~AA Osman, Dimitrios Tzionas, and Michael~J Black.
\newblock Expressive body capture: 3d hands, face, and body from a single image.
\newblock In \emph{CVPR}, 2019.

\bibitem[Petrovich et~al.(2021)Petrovich, Black, and Varol]{petrovich2021action}
Mathis Petrovich, Michael~J Black, and G{\"u}l Varol.
\newblock Action-conditioned 3d human motion synthesis with transformer vae.
\newblock In \emph{ICCV}, 2021.

\bibitem[Petrovich et~al.(2022)Petrovich, Black, and Varol]{petrovich2022temos}
Mathis Petrovich, Michael~J Black, and G{\"u}l Varol.
\newblock Temos: Generating diverse human motions from textual descriptions.
\newblock In \emph{ECCV}, 2022.

\bibitem[Plappert et~al.(2016)Plappert, Mandery, and Asfour]{plappert2016kit}
Matthias Plappert, Christian Mandery, and Tamim Asfour.
\newblock The kit motion-language dataset.
\newblock \emph{Big data}, 2016.

\bibitem[Preechakul et~al.(2022)Preechakul, Chatthee, Wizadwongsa, and Suwajanakorn]{preechakul2022diffusion_autoencoder}
Konpat Preechakul, Nattanat Chatthee, Suttisak Wizadwongsa, and Supasorn Suwajanakorn.
\newblock Diffusion autoencoders: Toward a meaningful and decodable representation.
\newblock In \emph{CVPR}, 2022.

\bibitem[Raab et~al.(2023)Raab, Leibovitch, Li, Aberman, Sorkine-Hornung, and Cohen-Or]{raab2023modi}
Sigal Raab, Inbal Leibovitch, Peizhuo Li, Kfir Aberman, Olga Sorkine-Hornung, and Daniel Cohen-Or.
\newblock Modi: Unconditional motion synthesis from diverse data.
\newblock In \emph{CVPR}, 2023.

\bibitem[Radford et~al.(2021)Radford, Kim, Hallacy, Ramesh, Goh, Agarwal, Sastry, Askell, Mishkin, Clark, et~al.]{radford2021learning_clip}
Alec Radford, Jong~Wook Kim, Chris Hallacy, Aditya Ramesh, Gabriel Goh, Sandhini Agarwal, Girish Sastry, Amanda Askell, Pamela Mishkin, Jack Clark, et~al.
\newblock Learning transferable visual models from natural language supervision.
\newblock In \emph{ICML}, 2021.

\bibitem[Raffel et~al.(2020)Raffel, Shazeer, Roberts, Lee, Narang, Matena, Zhou, Li, and Liu]{t5_raffel2020exploring}
Colin Raffel, Noam Shazeer, Adam Roberts, Katherine Lee, Sharan Narang, Michael Matena, Yanqi Zhou, Wei Li, and Peter~J Liu.
\newblock Exploring the limits of transfer learning with a unified text-to-text transformer.
\newblock \emph{JMLR}, 2020.

\bibitem[Rombach et~al.(2022)Rombach, Blattmann, Lorenz, Esser, and Ommer]{rombach2022high_latentdiffusion_ldm}
Robin Rombach, Andreas Blattmann, Dominik Lorenz, Patrick Esser, and Bj{\"o}rn Ommer.
\newblock High-resolution image synthesis with latent diffusion models.
\newblock In \emph{CVPR}, 2022.

\bibitem[Runge(1895)]{Runge}
Carl Runge.
\newblock {\"U}ber die numerische aufl{\"o}sung von differentialgleichungen.
\newblock \emph{Mathematische Annalen}, 1895.

\bibitem[Saharia et~al.(2022)Saharia, Chan, Saxena, Li, Whang, Denton, Ghasemipour, Ayan, Mahdavi, Lopes, et~al.]{saharia2022photorealistic_imagen}
Chitwan Saharia, William Chan, Saurabh Saxena, Lala Li, Jay Whang, Emily Denton, Seyed Kamyar~Seyed Ghasemipour, Burcu~Karagol Ayan, S~Sara Mahdavi, Rapha~Gontijo Lopes, et~al.
\newblock Photorealistic text-to-image diffusion models with deep language understanding.
\newblock In \emph{NeurIPS}, 2022.

\bibitem[Salimans and Ho(2022)]{salimans2022progressive}
Tim Salimans and Jonathan Ho.
\newblock Progressive distillation for fast sampling of diffusion models.
\newblock In \emph{ICLR}, 2022.

\bibitem[Sohl-Dickstein et~al.(2015)Sohl-Dickstein, Weiss, Maheswaranathan, and Ganguli]{sohl2015deep}
Jascha Sohl-Dickstein, Eric Weiss, Niru Maheswaranathan, and Surya Ganguli.
\newblock Deep unsupervised learning using nonequilibrium thermodynamics.
\newblock In \emph{ICML}, 2015.

\bibitem[Song et~al.(2021{\natexlab{a}})Song, Meng, and Ermon]{song2020denoising_ddim}
Jiaming Song, Chenlin Meng, and Stefano Ermon.
\newblock Denoising diffusion implicit models.
\newblock In \emph{ICLR}, 2021{\natexlab{a}}.

\bibitem[Song et~al.(2021{\natexlab{b}})Song, Sohl-Dickstein, Kingma, Kumar, Ermon, and Poole]{song2021scorebased_sde}
Yang Song, Jascha Sohl-Dickstein, Diederik~P Kingma, Abhishek Kumar, Stefano Ermon, and Ben Poole.
\newblock Score-based generative modeling through stochastic differential equations.
\newblock In \emph{ICLR}, 2021{\natexlab{b}}.

\bibitem[Song et~al.(2023)Song, Dhariwal, Chen, and Sutskever]{song2023consistency}
Yang Song, Prafulla Dhariwal, Mark Chen, and Ilya Sutskever.
\newblock Consistency models.
\newblock In \emph{ICML}, 2023.

\bibitem[Tevet et~al.(2023)Tevet, Raab, Gordon, Shafir, Cohen-or, and Bermano]{tevet2022human_mdm}
Guy Tevet, Sigal Raab, Brian Gordon, Yoni Shafir, Daniel Cohen-or, and Amit~Haim Bermano.
\newblock Human motion diffusion model.
\newblock In \emph{ICLR}, 2023.

\bibitem[Vaswani et~al.(2017)Vaswani, Shazeer, Parmar, Uszkoreit, Jones, Gomez, Kaiser, and Polosukhin]{vaswani2017attention}
Ashish Vaswani, Noam Shazeer, Niki Parmar, Jakob Uszkoreit, Llion Jones, Aidan~N Gomez, {\L}ukasz Kaiser, and Illia Polosukhin.
\newblock Attention is all you need.
\newblock \emph{NeurIPS}, 2017.

\bibitem[Wu et~al.(2023)Wu, Wang, Gong, Liu, Xiong, Ranjan, Krishnamoorthi, Chandra, and Liu]{wu2023fast}
Lemeng Wu, Dilin Wang, Chengyue Gong, Xingchao Liu, Yunyang Xiong, Rakesh Ranjan, Raghuraman Krishnamoorthi, Vikas Chandra, and Qiang Liu.
\newblock Fast point cloud generation with straight flows.
\newblock In \emph{CVPR}, 2023.

\bibitem[Yi et~al.(2023)Yi, Liang, Liu, Cao, Wen, Bolkart, Tao, and Black]{yi2023generating_talkshow}
Hongwei Yi, Hualin Liang, Yifei Liu, Qiong Cao, Yandong Wen, Timo Bolkart, Dacheng Tao, and Michael~J Black.
\newblock Generating holistic 3d human motion from speech.
\newblock In \emph{CVPR}, 2023.

\bibitem[Zanfir et~al.(2021)Zanfir, Zanfir, Bazavan, Freeman, Sukthankar, and Sminchisescu]{zanfir2021thundr}
Mihai Zanfir, Andrei Zanfir, Eduard~Gabriel Bazavan, William~T Freeman, Rahul Sukthankar, and Cristian Sminchisescu.
\newblock Thundr: Transformer-based 3d human reconstruction with markers.
\newblock In \emph{ICCV}, 2021.

\bibitem[Zhang et~al.(2023{\natexlab{a}})Zhang, Zhang, Cun, Huang, Zhang, Zhao, Lu, and Shen]{zhang2023generating_t2mgpt}
Jianrong Zhang, Yangsong Zhang, Xiaodong Cun, Shaoli Huang, Yong Zhang, Hongwei Zhao, Hongtao Lu, and Xi Shen.
\newblock T2m-gpt: Generating human motion from textual descriptions with discrete representations.
\newblock In \emph{CVPR}, 2023{\natexlab{a}}.

\bibitem[Zhang et~al.(2022)Zhang, Cai, Pan, Hong, Guo, Yang, and Liu]{zhang2022motiondiffuse}
Mingyuan Zhang, Zhongang Cai, Liang Pan, Fangzhou Hong, Xinying Guo, Lei Yang, and Ziwei Liu.
\newblock Motiondiffuse: Text-driven human motion generation with diffusion model.
\newblock In \emph{arXiv}, 2022.

\bibitem[Zhang et~al.(2023{\natexlab{b}})Zhang, Guo, Pan, Cai, Hong, Li, Yang, and Liu]{zhang2023remodiffuse}
Mingyuan Zhang, Xinying Guo, Liang Pan, Zhongang Cai, Fangzhou Hong, Huirong Li, Lei Yang, and Ziwei Liu.
\newblock Remodiffuse: Retrieval-augmented motion diffusion model.
\newblock In \emph{ICCV}, 2023{\natexlab{b}}.

\bibitem[Zhang et~al.(2020)Zhang, Hassan, Neumann, Black, and Tang]{zhang2020generating}
Yan Zhang, Mohamed Hassan, Heiko Neumann, Michael~J Black, and Siyu Tang.
\newblock Generating 3d people in scenes without people.
\newblock In \emph{CVPR}, 2020.

\bibitem[Zhang et~al.(2021)Zhang, Black, and Tang]{zhang2021we}
Yan Zhang, Michael~J Black, and Siyu Tang.
\newblock We are more than our joints: Predicting how 3d bodies move.
\newblock In \emph{CVPR}, 2021.

\bibitem[Zheng et~al.(2023)Zheng, Lu, Chen, and Zhu]{zheng2023improved}
Kaiwen Zheng, Cheng Lu, Jianfei Chen, and Jun Zhu.
\newblock Improved techniques for maximum likelihood estimation for diffusion odes.
\newblock In \emph{ICML}, 2023.

\bibitem[Zhu et~al.(2023)Zhu, Ma, Ro, Ci, Zhang, Shi, Gao, Tian, and Wang]{zhu2023human_motionsurvey23}
Wentao Zhu, Xiaoxuan Ma, Dongwoo Ro, Hai Ci, Jinlu Zhang, Jiaxin Shi, Feng Gao, Qi Tian, and Yizhou Wang.
\newblock Human motion generation: A survey.
\newblock In \emph{arXiv}, 2023.

\end{thebibliography}

}

% WARNING: do not forget to delete the supplementary pages from your submission 
\clearpage
\setcounter{page}{1}
\maketitlesupplementary
\onecolumn
\tableofcontents

\section{Societal Impact}
This work provides a powerful tool for quickly generating highly realistic human pose series. It has the potential to significantly expedite the process of creating human pose-related artwork, thus enabling the democratization of creativity. However, on the flip side, this tool also poses a challenge to the community in terms of creating safety mechanisms to prevent the misuse of generative models for deep-fake and misleading content creation with potentially controversial intentions.

\section{More results}
\label{sec:more_results}

\subsection{Demonstration Video}

\textcolor{red}{The video of our motion generation is attached in the supplementary files.}

\subsection{Qualitative Results}

%\paragraph{Straight Trajectory.} ~\Cref{fig:trajectory_x1_est_10steps}.

%~\Cref{fig:trajectory_curve}.

\begin{figure}
    \centering
  \includegraphics[width=.6\linewidth]{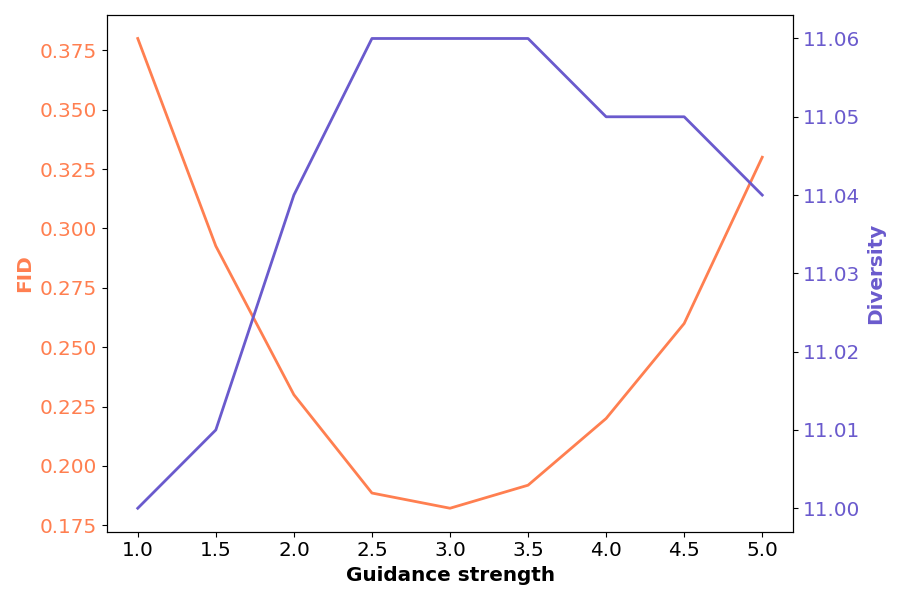}
  \captionof{figure}{\textbf{The FID and Diversity on the guidance strength} of our method on KIT-ML dataset.}
  \label{fig:guidance_strength}
\end{figure}

\paragraph{Guidance Strength.} We also explore the guidance strength, denoted as $s$, in the classifier-free guidance, as depicted in~\Cref{fig:guidance_strength} on dataset KIT-ML. Interestingly, we observe that a value of $s=3.0$ emerges as the optimal point, aligning with the findings reported in MDM~\citep{tevet2022human_mdm}, which further validates the consistency of our results with prior research. 

\begin{figure}
    \centering
  \includegraphics[width=.99\linewidth]{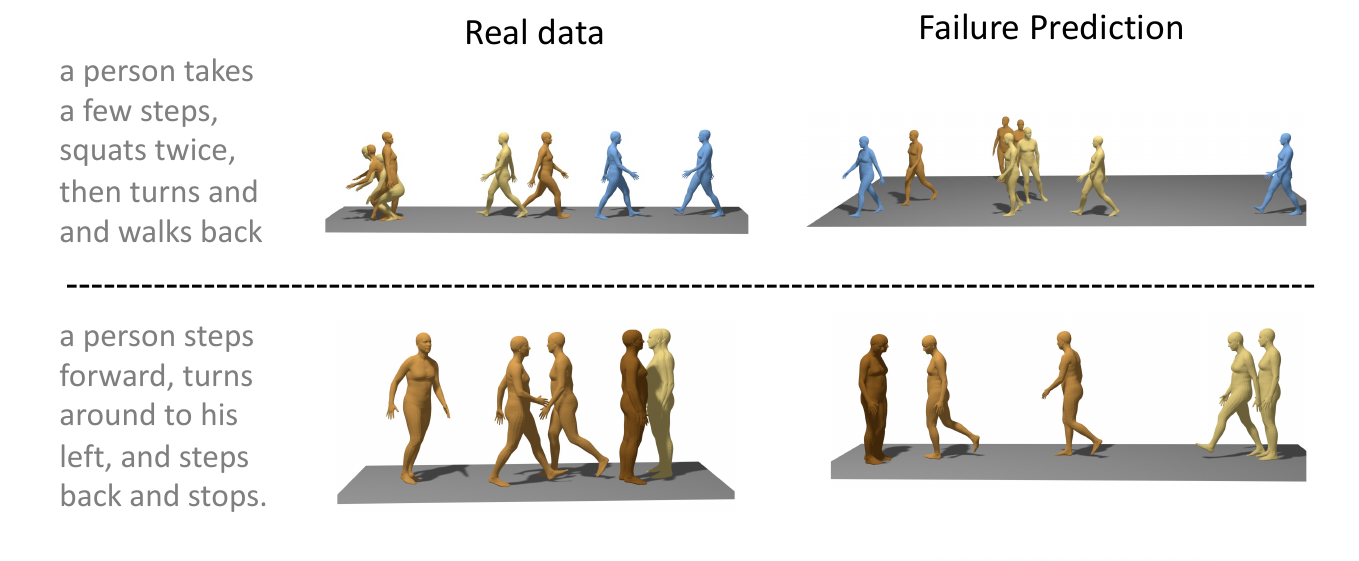}
  \captionof{figure}{\textbf{Failure case visualization} of motion in-between editing and text-to-motion synthesis on HumanML3D dataset.}
  \label{fig:failure_case}
\end{figure}

\paragraph{Failure Case.}  We have indeed observed some typical failure cases, as illustrated in~\Cref{fig:failure_case}. In the first example, the model struggles to effectively interpret the prompt related to a "squats twice" motion, resulting in an interruption in the ground truth motion. This is attributed to the rareness of certain words or concepts, which the model may not consistently capture, affecting the alignment between the prompt and motion.
In the second example, it appears that the model did not effectively interpret the hint from the prompt to turn around and step back, which led to the observed result. We have also observed failure cases in MDM~\citep{tevet2022human_mdm} that are comparable or even worse using the same prompt. This underscores that interpreting multiple fine-grained textual descriptions into motion remains a formidable challenge. 

%The second example pertains to an issue known as the ``upside-down problem", which is observed in MDM~\cite{tevet2022human_mdm} as well. This problem is well-documented in the community, as mentioned \href{https://github.com/GuyTevet/motion-diffusion-model/issues/101#issuecomment-1459588236}{in this GitHub issue}. To mitigate this issue further, we may consider incorporating a foot contact loss.

%\wy{Also, we observe the shaking problem in some generated cases. This could attributed to ? }

\paragraph{More qualitative results about the text-to-motion synthesis}

\begin{figure}
    \centering
    \includegraphics[width=1\textwidth]{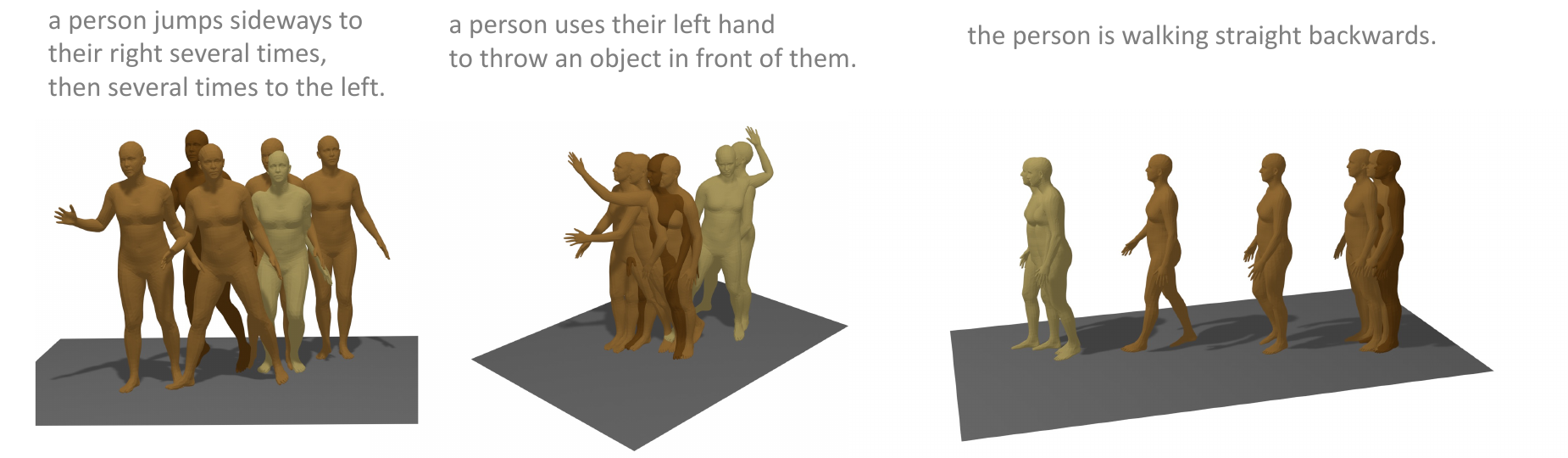}
    \caption{\textbf{More visualization of text-to-motion synthesis on HumanML3D dataset}. The gradient of colors, ranging from light to dark, signifies the passage of time. }
    \label{fig:text2motion}
\end{figure}

We demonstrate more visualization of the text-to-motion synthesis in~\Cref{fig:text2motion}. It shows that the synthesized motion aligns with the prompt textual description.

\begin{figure}
    \centering
    \includegraphics[width=1\textwidth]{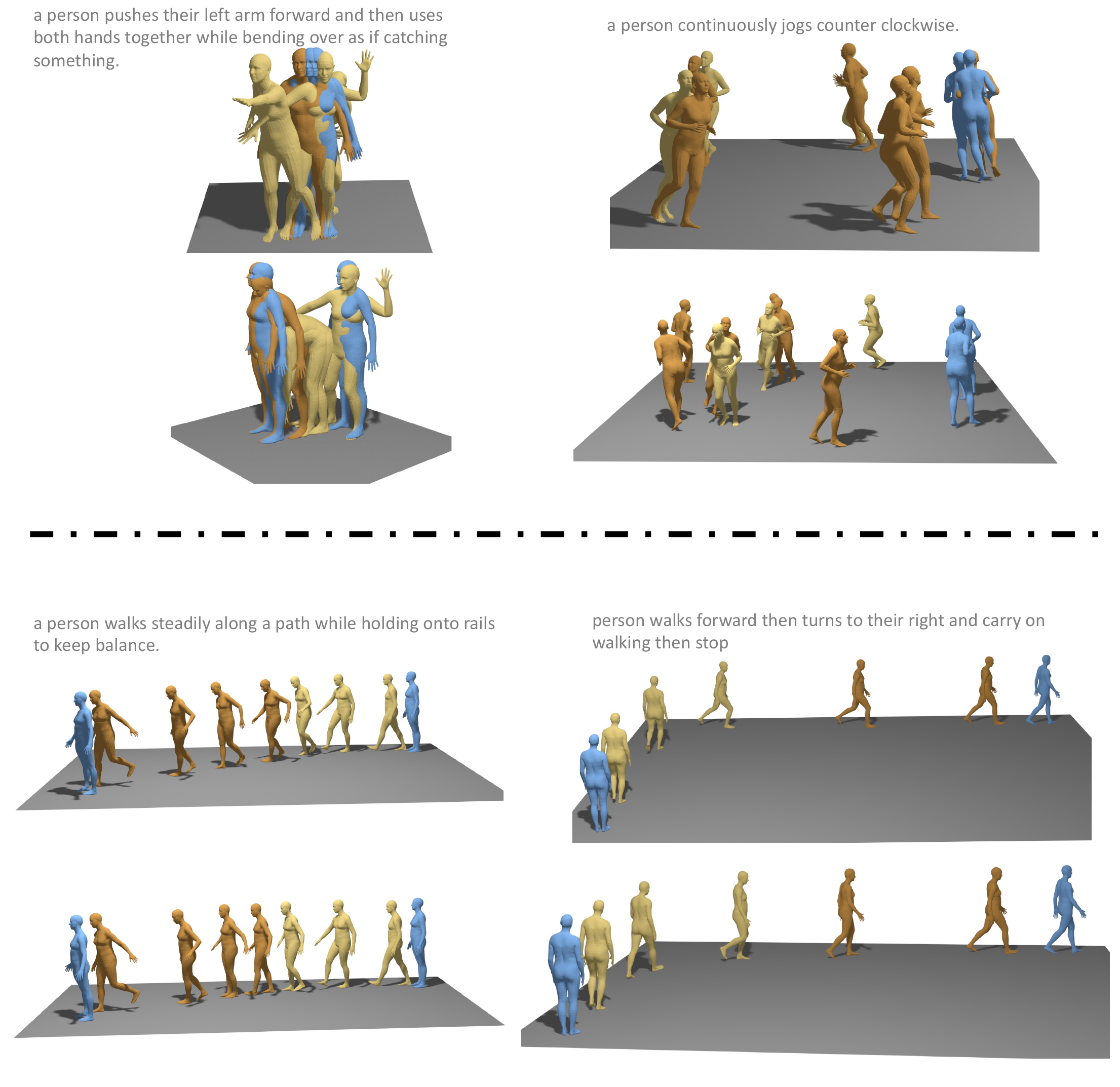}
    \caption{\textbf{More results about sampling trajectory rewriting in in-between editing.} Light blue frames denote motion input, while bronze frames signify generated motion. The gradient of colors, ranging from light to dark, signifies the passage of time. }
    \label{fig:in_between_more}
\end{figure}

\paragraph{More qualitative results about in-between.} We present our visualization of in-between in~\Cref{fig:in_between_more}. Given the prefix and suffix poses as well as the prompt, our model can generate reasonable motions in between.

\subsection{Quantitative Results}
\label{supp:extra_quantitative}

\begin{comment}
%\lipsum[1-4]
\begin{wrapfigure}{r}{6cm}
  \centering
  
   \includegraphics[width=0.4\textwidth]{fig/sampling_speed_compare_humanml3d.png}
    \caption{\textbf{FID vs Number of Forward Evaluation} on HumanML-3D. Our method can achieve better and fewer sampling steps at the same timestep.}
    \label{fig:sampling_speed_compare_humanml3d}
  %\rule{3cm}{7cm}
  \vspace{-20pt} % This removes the white box on the second page
\end{wrapfigure}
%\lipsum[1-6]
\end{comment}

\paragraph{Result on action-to-motion dataset HumanAct12.} We present our results on the action-to-motion dataset HumanAct12 in~\Cref{tab:humanact12}. Our method achieves results that are comparable to various baselines, all while significantly reducing the number of forward evaluations and utilizing fewer parameters.

\paragraph{The choice of the text encoder.} We also explored the possibility of replacing the default CLIP encoder with a more advanced encoder, such as T5~\citep{t5_raffel2020exploring} in \Cref{tab:text_encoder}. We did not observe an obvious gain in comparison with CLIP embedding. Interestingly, our findings sharply contrast with the conclusions drawn in Imagen~\citep{saharia2022photorealistic_imagen}. This aligns with the results reported in MLD~\citep{chen2023executing_mld} (as seen in Table 7 in the Appendix). The disparity in outcomes could potentially be attributed to the current scale of the HumanML dataset, which may not be extensive enough to fully leverage the capabilities of a more sophisticated text encoder like T5.

\begin{table}
\centering
\caption{\textbf{Evaluation of action-to-motion on the HumanAct12 dataset. } NFE denotes the number of function evaluations. $\rightarrow$ indicates that closer to real is better.}
%Our model leads the board in three out of four metrics. Ground-truth evaluation results are slightly different for each of the works, due to implementation differences, such as python package versions. It is important to assess the diversity and multimodality of each model using its own ground-truth results, as they are measured by their distance from GT. We show the GT metrics measured by our model and by the leading compared work, INR~\citep{cervantes2022implicit}.\textbf{Bold} indicates best result, $\underline{\text{underline}}$ indicates second best, $\pm$ indicates 95\% confidence interval, $\rightarrow$ indicates that closer to real is better.

\resizebox{.95\textwidth}{!}{
    \begin{tabular}{lcccccl}
    \toprule
    Method & NFE$\downarrow$ & $\text{FID}_{train}$ $\downarrow$ &
    Accuracy$\uparrow$ &
   Diversity$\rightarrow$ &
  Multimodality$\rightarrow$ &
   \#param $\downarrow$ \\

    \midrule
    %Real (INR) & $0.020^{\pm.010}$ & $0.997^{\pm.001}$ & $6.850^{\pm.050}$ & $2.450^{\pm.040}$\\ 

    {Real Motion}  & & $0.050^{\pm.000}$ & $0.990^{\pm.000}$ & $6.880^{\pm.020}$ & $2.590^{\pm.010}$\\   %

    \midrule
    
    Action2Motion~\citeyearpar{guo2020action2motion} & & $0.338^{\pm.015}$ & $0.917^{\pm.003}$ & $6.879^{\pm.066}$ & ${2.511^{\pm.023}}$\\ 
    
    ACTOR~\citeyearpar{petrovich2021action} & & $0.120^{\pm.000}$ &$0.955^{\pm.008}$ &${6.840^{\pm.030}}$ &$2.530^{\pm.020}$ \\ 

    INR~\citeyearpar{cervantes2022implicit} & &  ${0.088^{\pm.004}}$ & ${0.973^{\pm.001}}$ & $6.881^{\pm.048}$ & $2.569^{\pm.040}$ \\ 

    \midrule
    
    MDM~\cite{tevet2022human_mdm} & {1,000}& 
        $0.100^{\pm.000}$ & ${0.990^{\pm.000}}$ & ${6.860^{\pm.050}}$ & $2.520^{\pm.010}$ &23M\\
    MLD~\cite{chen2023executing_mld} &50& 
        $0.077^{\pm.004}$ & ${0.964^{\pm.002}}$ & ${6.831^{\pm.050}}$ & $2.824^{\pm.038}$& 26.9M  \\
    \midrule
    
    %\quad\quad w/o foot contact  & $\mathbf{0.080^{\pm.000}}$ & $\mathbf{0.990^{\pm.000}}$ & $6.810^{\pm.010}$ & $\mathbf{2.580^{\pm.010}}$ \\  %

     \textbf{Our MFM} &{10}& 
        $0.110^{\pm.000}$ & ${0.978^{\pm.000}}$ & ${6.850^{\pm.020}}$ & ${2.610}^{\pm.012}$ & 17.6M\\

    \bottomrule
    \end{tabular}
}
\label{tab:humanact12}
\end{table}

\begin{table}
    \centering
     \caption{\textbf{Ablation study about the text encoder on KIT-ML~\citep{plappert2016kit} test set.} $\rightarrow$ indicates that closer to real is better.}
    \begin{tabular}{c|cccc}
        \toprule
        \textbf{Encoder} & {FID $\downarrow$} & {MM-Dist $\downarrow$} & {Diversity $\rightarrow$} & {MModality $\uparrow$}\\
        \midrule
          {Real motion} & \et{0.002}{.000} & \et{2.974}{.008} & \et{9.503}{.065} & -  \\
         \midrule
        CLIP &  \et{0.1822}{.029} & \et{9.04}{.043} & \et{11.06}{.108} & \et{1.53}{.079} \\
        T5~\cite{t5_raffel2020exploring} & \et{0.2301}{.022} & \et{8.97}{.035} & \et{10.951}{.101} & \et{1.52}{.090} \\
         \bottomrule
    \end{tabular}
    \label{tab:text_encoder}
\end{table}

\section{Implementation Detail}
\label{sup:impl_detail}

\paragraph{Flow matching framework.}

For the sake of completeness, we reiterate the algorithm outlined in~\Cref{alg:flow_matching_pseudo}.

\begin{algorithm}
        \caption{Flow Matching Algorithm.}
 \label{alg:flow_matching_pseudo}
\begin{algorithmic}[1]
\STATE{\textbf{Input:} Empirical distribution $q_1$, gaussian distribution $q_0$,  batchsize $b$, initial network $v_{\theta}$.}
%\STATE{\textbf{Output:} An edited image $\mathbf{x}_{1}$.} 

\WHILE{Training}
 \STATE{Sample batches of size $b$ \textit{i.i.d.} from the datasets}
 \STATE{$\vx_0 \sim q_0(\vx_0); \quad \vx_1 \sim q_1(\vx_1)$}
\STATE {$t \sim \mathcal{U}(0, 1)$}
\STATE{\textcolor{blue}{\texttt{\# Interpolation.}}}
\STATE {$x_t \gets t \vx_1 + (1 - t) \vx_0$}

\STATE {$\text{FM}(\theta) \gets \| v_\theta(x_t,t) - (\vx_1 - \vx_0)\|^2$}
\STATE {$\theta \gets \mathrm{Update}(\theta, \nabla_\theta \text{FM}(\theta))$}
 \ENDWHILE
\STATE {\textbf{Return} $\text{FM}$}

\end{algorithmic}
\end{algorithm}

 \begin{algorithm}[H]
 \small
 \caption{%\small
 Euler Sampling algorithm with sampling trajectory rewriting.
 }
 \label{alg:traj_rewrite_python}
 \definecolor{codeblue}{rgb}{0.25,0.5,0.5}
 \definecolor{codekw}{rgb}{0.85, 0.18, 0.50}
 \lstset{
   backgroundcolor=\color{white},
   basicstyle=\fontsize{9.2pt}{9.2pt}\ttfamily\selectfont,
   columns=fullflexible,
   breaklines=true,
   captionpos=b,
   commentstyle=\fontsize{9.2pt}{9.2pt}\color{codeblue},
   keywordstyle=\fontsize{9.2pt}{9.2pt}\color{codekw},
   escapechar={|}, 
   xleftmargin=.02\textwidth, xrightmargin=.02\textwidth
 }
 \begin{lstlisting}[language=python]
 def trajectory_rewriting(model, noise, x1, mask,  edit_till = 0.2):
   # x1: origin data.
   # mask: 1 denotes known, 0 denotes unknown.
   # model: pretrained vector field predictor
   z = noise.detach().clone()
   dt = 1.0 / N
   est, traj = [], []

   for i in range(0, N, 1):  # fix-step Euler ODE solver
        t = i / N
        |\color{blue}x\_interp = noise * (1 - t) + x1 * t| # interpolate known area.
        if t <= edit_till:
             |\color{blue}z = z * (1 - mask) +  x\_interp * mask| # traj rewriting.
         pred = model(z, t) # vector field prediction
         pred = pred.detach().clone()
         _est_now = z + (1 - t)*pred
         est.append(_est_now)

         z = z.detach().clone() + pred * dt
         traj.append(z.detach().clone())

     return traj[-1], est
 \end{lstlisting}
 \end{algorithm}
 %\vspace{-18pt}

\paragraph{Transformer structure.}

\begin{figure}
    \centering
    \includegraphics[width=0.9\textwidth]{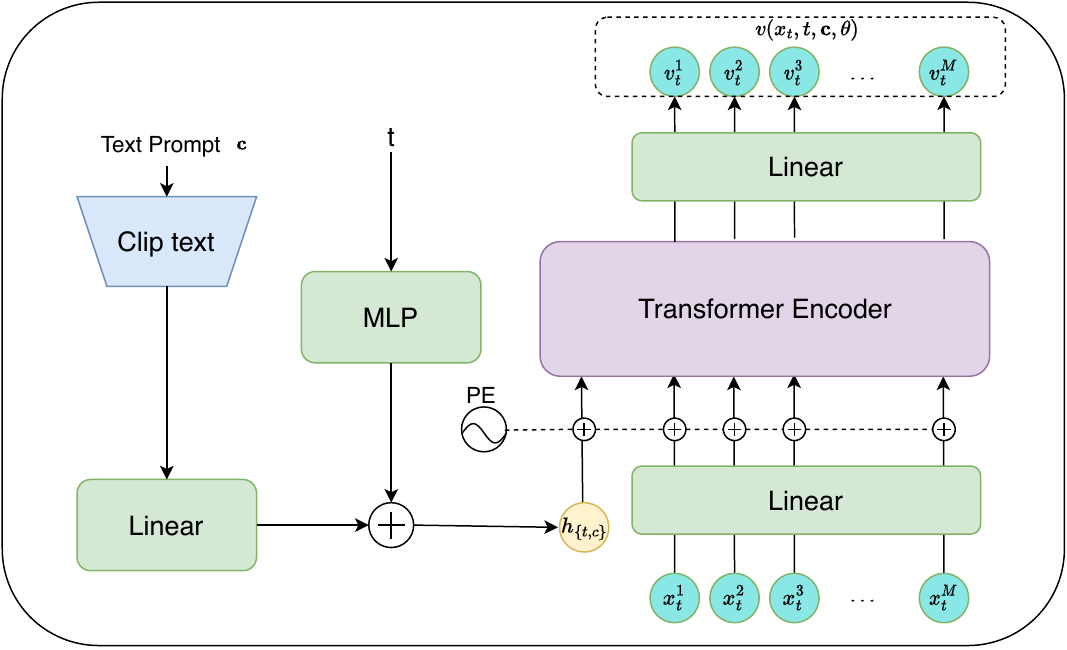}
    \caption{\textbf{The dynamic of the flow matching: transformer architecture of text-to-motion generation.} To model the temporal consistency of motion, we employ a transformer architecture. Our input comprises a text prompt $\bc$, timestep $t$, and noisy data $\bx_t$, which are combined as $\bh_{[\bx_t, \bc, t]}$. The output of our model is an estimated vector field denoted as $v(\bx_t,t,\bc;\theta)$, calculated at the position $\bx_t$ and timestep $t$. PE represents the position encoding used in our model. Our text-to-motion generation is conditioned on CLIP in a classifier-free manner~\citep{ho2021classifier}.}
    \label{fig:transformer}
\end{figure}

We illustrate the network structure of our model in~\Cref{fig:transformer}. We use CLIP~\citep{radford2021learning_clip} to encode the text prompt. Given the text embedding $\bc$, time embedding $t$, and noisy data $x_t$ with position encoding (PE), the transformer is able to ensure the temporal coherence of motion. The resulting output from our model is a vector field, symbolized by $v(\bx_t,t,\bc;\theta)$, determined at the position $\bx_t$ and the timestep $t$. Additionally, our text-to-motion generation is conditioned on CLIP in a classifier-free manner~\citep{ho2021classifier}.

\paragraph{Python pseudo code.} We provide the pseudo code for sampling trajectory rewriting in ~\Cref{alg:traj_rewrite_python}.

%\tao{why don't consider foot contat loss?}

%The followings are the hyperparameters and model details for all of our experiments.

\subsection{Training}

As the default text encoder, we employ CLIP~\citep{radford2021learning_clip}, and our flow matching model comprises 8 layers in both the transformer encoder. The feed-forward networks have an output dimensionality of $d_\text{ff} = 1024$, while the attention mechanisms employ an inner dimensionality of $d_\text{kv} = 512$ with 4 heads.
For training, all our models use the AdamW optimizer. The motion tokenizers are trained with a learning rate of $10^{-4}$ and a mini-batch size of 256. Training is conducted on 4 Tesla A5000 GPUs, taking approximately 3 days for HumanML3D and 2 days for KIT-ML.

In the case of text-to-motion tasks, we encode the textual prompts into $\bc$ utilizing a text encoder. The selection of the encoder may be based on preference. Conversely, for action-to-motion tasks, we employ pre-learned embeddings tailored to each distinct class $\bc$.

\subsection{Editing}
For sampling trajectory rewriting, we opt for a slightly larger value of $N=30$, but restrict the editing to the initial $t=0.2$ timesteps, effectively modifying the first 6 timesteps only.

\subsection{Dataset}

\paragraph{KIT Motion-Language (KIT-ML).} KIT-ML~\citep{plappert2016kit} contains 3,911 human motion sequences and 6{,}278 textual annotations. The total vocabulary size, that is the number of unique words disregarding capitalization and punctuation, is 1,623. Motion sequences are selected from KIT~\citep{mandery2015kit} and CMU datasets but downsampled into 12.5 frame-per-second (FPS). Each motion sequence is described by from 1 to 4 sentences. The average length of descriptions is approximately 8. Following~\cite{guo2022generating,guo2022tm2t}, the dataset is split into training, validation, and test sets with proportions of 80\%, 5\%, and 15\%, respectively. We select the model that achieves the best FID on the validation set and reports its performance on the test set.

\paragraph{HumanML3D.} HumanML3D~\citep{guo2022generating} is currently the largest 3D human motion dataset with textual descriptions. The dataset contains 14,616 human motion and 44{,}970 text descriptions. The entire textual descriptions are composed of  5{,}371 distinct words. The motion sequences are originally from AMASS~\citep{mahmood2019amass} and HumanAct12~\citep{guo2020action2motion} but with specific pre-processing: motion is scaled to 20 FPS; those that are longer than 10 seconds are randomly cropped to 10-second ones; they are then re-targeted to a default human skeletal template and properly rotated to face Z+ direction initially. Each motion is paired with at least 3 precise textual descriptions. The average length of descriptions is approximately 12. According to~\cite{guo2022generating}, the dataset is split into training, validation, and test sets with proportions of 80\%, 5\%, and 15\%, respectively. We select the model that achieves the best FID on the validation set and reports its performance on the test set.

\paragraph{HumanAct12.} To further validate the effectiveness of our method, we also test it on the action-to-motion generation task. Action-to-motion involves generating motion sequences based on input action classes, typically represented as scalar values. We employ the well-established benchmark dataset, HumanAct12~\citep{guo2020action2motion}, which serves as the standard evaluation benchmark for action-to-motion models. We evaluate our model using the set of metrics suggested by Guo et al. (2020), namely the Fréchet Inception Distance (FID), action recognition accuracy, diversity, and multimodality. The combination of these metrics constitutes a comprehensive measure of the realism and diversity of the generated motions.

HumanAct12~\citep{guo2020action2motion} offers approximately 1200 motion clips, organized into 12 action categories, with 47 to 218 samples per label. We adhere to the cross-subject testing protocol used by current works, with 225-345 samples per action class.
For both datasets we use the sequences provided by \citet{petrovich2021action}.

In the case of action-to-motion, the only change would be the substitution of the text embedding by an action embedding. Since action is represented by a scalar, its embedding is fairly simple; each input action class scalar is converted into a learned embedding of the transformer dimension.

Consistent with~\cite{guo2022generating}, the maximum motion sequence length on both datasets is 196, while the minimum lengths are 40 for HumanML3D~\citep{guo2022generating} and 24 for KIT-ML~\citep{plappert2016kit}.
For both the HumanML3D~\citep{guo2022generating} and KIT-ML~\citep{plappert2016kit} datasets, the motion sequences are truncated to $T=60$ during training.

\section{More details on the evaluation metrics and the motion representations.}
\label{sec:supp_metrics}

\subsection{Evaluation metrics}

To enhance clarity and comprehensibility for our readers, we reiterate the descriptions of the relevant metrics here to prevent any potential confusion.

We employ standard metrics as defined in~\cite{guo2022generating} and~\cite{tevet2022human_mdm} for our evaluations. To ensure robustness, we repeat each evaluation 20 times and report the average values along with 95\% confidence intervals. 

We detail the calculation of several evaluation metrics, which are proposed in ~\cite{guo2022generating}. We denote ground-truth motion features, generated motion features, and text features as $f_{gt}$, $f_{pred}$, and $f_{text}$. Note that these features are extracted with pretrained networks in~\cite{guo2022generating}.

\paragraph{FID.}
FID is widely used to evaluate the overall quality of the generation. We obtain FID by
\begin{equation}
\text{FID} = \lVert \mu_{gt} - \mu_{pred}\rVert^2 - \text{Tr}(\Sigma_{gt} + \Sigma_{pred} - 2(\Sigma_{gt}\Sigma_{pred})^{\frac{1}{2}})
\label{formula:fid}
\end{equation}
where $\mu_{gt}$ and $\mu_{pred}$ are mean of $f_{gt}$ and $f_{pred}$. $\Sigma$ is the covariance matrix and $\text{Tr}$ denotes the trace of a matrix.

\paragraph{MM-Dist.}
MM-Dist measures the distance between the text embedding and the generated motion feature. Given N randomly generated samples, the MM-Dist measures the feature-level distance between the motion and the text. Precisely, it computes the average Euclidean distances between each text feature and the
generated motion feature from this text:
\begin{equation}
\text{MM-Dist} = \frac{1}{N}\sum_{i=1}^{N}\lVert f_{pred,i} - f_{text,i}\rVert
\label{formula:mm-dis}
\end{equation}
where $f_{pred,i}$ and  $f_{text,i}$ are the features of the i-th text-motion pair. 

\paragraph{Diversity.} Diversity measures the variance of the whole motion sequences across the dataset. We randomly sample $S_{dis}$ pairs of motion and each pair of motion features is denoted by $f_{pred,i}$ and $f_{pred,i}'$. The diversity can be calculated by
\begin{equation}
\text{Diversity} = \frac{1}{S_{dis}}\sum_{i=1}^{S_{dis}}||f_{pred,i} - f_{pred,i}'||
\label{formula:diversity}
\end{equation}
In our experiments, we set $S_{dis}$ to 300 as \cite{guo2022generating}.

\paragraph{MModality.} MModality measures the diversity of human motion generated from the same text description. Precisely, for the i-th text description, we generate motion 30 times and then sample two subsets containing 10 motions. We denote features of the j-th pair of the i-th text description by ($f_{pred,i,j}$, $f_{pred,i,j}'$). The MModality is defined as follows:
\begin{equation}
\text{MModality} = \frac{1}{10N}\sum_{i=1}^{N}\sum_{j=1}^{10}\lVert f_{pred,i,j} - f_{pred,i,j}'\rVert
\label{formula:mmodality}
\end{equation}

\paragraph{Multimodal Distance.} We calculate the multimodal distance as the average Euclidean distance between the motion feature of each generated motion and the text feature of its corresponding description in the test set. A lower value implies better multimodal distance.

\paragraph{R Precision. (top-3)} For each generated motion, R Precision assesses how relevant the generated motions are to the given input prompts. Based on the ground-truth text, a batch of mismatched descriptions is randomly selected from the test set. We calculate the Euclidean distance between the motion feature and text feature of each description in the pool. We count the average accuracy at the top 3 places. If the ground truth entry falls into the top 3 candidates, we treat it as True Positive retrieval. We use a batch size of 32 (i.e. 31 negative examples).

\subsection{Motion representation}
\label{sup:representation}

We use the same motion representation as ~\cite{guo2022generating}. Each pose is represented by $(\dot{r}^a, \dot{r}^x, \dot{r}^z, r^y, j^p, j^v, j^r, c^f)$, where $\dot{r}^a \in \mathbb{R}$ is the global root angular velocity along the Y-axis;  $\dot{r}^x \in \mathbb{R}, \dot{r}^z \in \mathbb{R}$ are the global root velocity in the X-Z plane; $j^p \in \mathbb{R}^{3j}, j^v \in \mathbb{R}^{3j}, j^r \in \mathbb{R}^{6j}$ are the local pose positions, velocity and rotation with $j$ the number of joints; $c^f\in \mathbb{R}^4$ is the foot contact features calculated by the heel and toe joint velocity.

%In terms of dataset specifications, the motion dimension for HumanML stands at $T\times$263, while for KIT, it amounts to $T\times$251, and the $T$ denotes the temporal number. This variation arises from the distinct number of joints involved. For a more comprehensive explanation of how the vector is constructed, you can refer to the implementation detailed in ~\cite{guo2022generating}.

\end{document}